\documentclass{article} 
\usepackage{nips14submit_e,times}
\usepackage{hyperref}
\usepackage{url}
\usepackage{multirow}

\usepackage{framed}
\usepackage{algorithmic}
\usepackage{graphicx} 
\usepackage{amssymb}
\usepackage{amsmath}
\usepackage{amsthm}
\usepackage{caption}
\usepackage{subcaption}
\usepackage{listings}
\usepackage{pdfpages}

\usepackage[compact]{titlesec}
\titlespacing{\section}{0pt}{0.5ex}{0.3ex}
\titlespacing{\subsection}{0pt}{0.2ex}{0ex}
\titlespacing{\subsubsection}{0pt}{0.1ex}{0ex}

\usepackage{paralist}

\makeatletter
\ifcase \@ptsize \relax
  \newcommand{\miniscule}{\@setfontsize\miniscule{4}{5}}
\or
  \newcommand{\miniscule}{\@setfontsize\miniscule{5}{6}}
\or
  \newcommand{\miniscule}{\@setfontsize\miniscule{5}{6}}
\fi
\makeatother

\newcommand {\aplt} {\ {\raise-.5ex\hbox{$\buildrel<\over\sim$}}\ }

\newcommand{\eqn}[1]{Eqn.~\ref{eqn:#1}}
\newcommand{\fig}[1]{Fig.~\ref{fig:#1}}
\newcommand{\tab}[1]{Table~\ref{tab:#1}}
\newcommand{\secc}[1]{Section~\ref{sec:#1}}
\def\etal{{\textit{et~al.~}}}
\newcommand{\BigO}[1]{\ensuremath{\operatorname{O}\left(#1\right)}}
\usepackage[symbol*]{footmisc}

\DefineFNsymbolsTM{myfnsymbols}{
  \textasteriskcentered *
  \textdagger    \dagger
  \textdaggerdbl \ddagger
  \textsection   \mathsection
  \textbardbl    \|%
  \textparagraph \mathparagraph
}%

\title{Learning to Discover\\ Efficient Mathematical Identities}

\author{
Wojciech Zaremba\\
Dept. of Computer Science\\
Courant Institute\\
New York Unviersity
\And
Karol Kurach\\
Google Zurich \& \\
Dept. of Computer Science\\
University of Warsaw
\And
Rob Fergus \\
Dept. of Computer Science\\
Courant Institute\\
New York Unviersity
}

\nipsfinalcopy 

\begin{document}

\maketitle

\vspace{-5mm}
\begin{abstract}
  In this paper we explore how machine learning techniques can be
  applied to the discovery of efficient mathematical identities. We
  introduce an attribute grammar framework for representing symbolic
  expressions. Given a grammar of math operators, we build trees that
  combine them in different ways, looking for compositions that are analytically equivalent to a target
  expression but of lower computational complexity. However, as the
  space of trees grows exponentially with the complexity of the
  target expression, brute force search is impractical for all but the
  simplest of expressions. Consequently, we introduce two novel
  learning approaches that are able to learn from simpler expressions
  to guide the tree search. The first of these is a simple $n$-gram
  model, the other being a recursive neural-network. We show how
  these approaches enable us to derive complex identities, beyond
  reach of brute-force search, or human derivation.
\end{abstract}

\vspace{-2mm}
\section{Introduction}
\vspace{-2mm} 
Machine learning approaches have proven highly effective for
statistical pattern recognition problems, such as those encountered in
speech or vision. However, their use in symbolic settings has been
limited. In this paper, we explore how learning can be applied to the
discovery of mathematical identities. Specifically, we propose
methods for finding computationally efficient versions of a given target
expression. That is, finding a new expression which computes an identical
result to the target, but has a lower complexity (in time and/or
space).

We introduce a framework based on attribute grammars
\cite{knuth1968semantics} that allows symbolic expressions to be
expressed as a sequence of grammar rules. 
Brute-force enumeration of all valid rule combinations
allows us to discover efficient versions of the target, including
those too intricate to be discovered by human manipulation. But
for complex target expressions this strategy quickly
becomes intractable, due to the exponential number of combinations that
must be explored. In practice, a random search within the grammar tree
is used to avoid memory problems, but the chance of finding a matching
solution becomes vanishingly small for complex targets.

To overcome this limitation, we use machine learning to produce
a search strategy for the grammar trees that selectively explores 
branches likely (under the model) to yield a solution. The training data for
the model comes from solutions discovered for simpler target
expressions. We investigate several different learning approaches. The
first group are $n$-gram models, which learn pairs, triples etc. of
expressions that were part of previously discovered solutions, thus
hopefully might be part of the solution for the current target. We
also train a recursive neural network (RNN) that operates
within the grammar trees. This model is first pretrained to learn a continuous representation for symbolic
expressions. Then, using this representation we learn to predict the
next grammar rule to add to the current expression to yield an
efficient version of the target. 

Through the use of learning, we are able to dramatically widen the
complexity and scope of expressions that can be handled in our
framework. We show examples of (i) $\BigO{n^3}$ target expressions
which can be computed in $\BigO{n^2}$ time (e.g.~see Examples 1 \& 2), and
(ii) cases where naive evaluation of the target would require {\em
  exponential} time, but can be computed in $\BigO{n^2}$ or
$\BigO{n^3}$ time. The majority of these examples are too complex to
be found manually or by exhaustive search and, as far as we are aware,
are previously undiscovered. All code and evaluation data can be found at \url{https://github.com/kkurach/math_learning}.

In summary our contributions are:
\begin{compactitem}
\item A novel grammar framework for finding
  efficient versions of symbolic expressions.
\item Showing how machine learning techniques can be integrated into
  this framework, and demonstrating how training models on
  simpler expressions can help which the discovery of more complex ones.
\item A novel application of a recursive neural-network to learn a continuous representation of 
  mathematical structures, making the symbolic domain accessible to
  many other learning approaches.
\item The discovery of many new mathematical identities which offer a significant reduction
  in computational complexity for certain expressions.
\end{compactitem}

\begin{minipage}{\linewidth}
\begin{framed}
\begin{flushleft}
\vspace{0mm}
{\bf Example 1:} Assume we are given matrices $A \in \mathbb{R}^{n \times m}$, $B \in
\mathbb{R}^{m \times p}$. 
 We wish to compute the target expression: \texttt{sum(sum(A*B))}, i.e.~: 
$\sum_{n,p} AB = \sum_{i = 1}^n \sum_{j = 1}^m \sum_{k = 1}^p A_{i, j} B_{j, k} $
which naively takes $O(nmp)$ time. Our framework is able to discover
an efficient version of the formula, that computes the same result in $O(n(m+p))$
time: {\small \texttt{sum((sum(A, 1) * B)', 1)}}. \\
Our framework builds {\em grammar trees} that explore valid compositions of expressions from the
  grammar, using a {\em search strategy}.  In this example, the naive
  strategy of randomly choosing permissible rules suffices and we can
  find another tree which matches the target expression in reasonable
  time. Below, we show trees for (i) the original expression and (ii) the efficient
formula which avoids the use of a matrix-matrix multiply operation, hence is
efficient to compute.
\vspace{-2mm} 
\begin{center}
\includegraphics[width=0.45\linewidth]{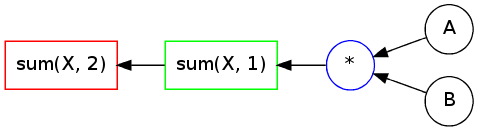}
\quad
\quad
\includegraphics[width=0.45\linewidth]{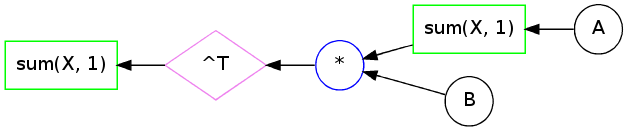}
\end{center}
\vspace{-2mm}
\end{flushleft}
\label{fig:example_ab}

\vspace{-3mm}

---------------------

{\bf Example 2:} Consider the target expression:
\texttt{sum(sum((A*B)$^k$))}, where $k=6$. For an expression of this degree,
there 
are 9785 possible grammar trees and the naive strategy used in Example
1 breaks down. We therefore {\em learn} a search strategy, training a model on successful trees
from simpler expressions, such as those for $k=2,3,4,5$. Our learning approaches
capture the common structure within the solutions, evident below, so can find an efficient $\BigO{nm}$
expression for this target: \\
{\scriptsize
 $k=2$: \texttt{sum((((((sum(A, 1)) * B) * A) * B)'), 1)}\\
$k=3$: \texttt{sum((((((((sum(A, 1)) * B) * A) * B) * A) * B)'), 1)}\\
$k=4$: \texttt{sum((((((((((sum(A, 1)) * B) * A) * B) * A) * B) * A) * B)'), 1)}\\
$k=5$: \texttt{sum((((((((((((sum(A, 1)) * B) * A) * B) * A) * B) * A)  * B) * A) * B)'), 1)}\\
$k=6$: \texttt{sum(((((((((((((sum(A, 1) * B) * A) * B) *A) * B) * A) * B)* A) * B) * A) * B)'), 1)}}
\end{framed}

\end{minipage}

\subsection{Related work}\label{relatedwork}

The problem addressed in this paper overlaps with the areas of theorem
proving \cite{chang1973symbolic,cook1971complexity,fitting1996first},
program induction \cite{nordin1997evolutionary, wong1997evolutionary}
and probabilistic programming
\cite{goodman2012church,pfeffer2011practical}.  These domains involve
the challenging issues of undecidability, the halting problem, and a
massive space of potential computation. However, we limit our domain
to computation of polynomials with fixed degree $k$, where
undecidability and the halting problem are not present, and the space
of computation is manageable (i.e.~it grows exponentially, but not
super-exponentially).  Symbolic computation engines, such as Maple
\cite{char1991maple} and Mathematica \cite{wolfram1996mathematica} are
capable of simplifying expressions by collecting terms but do not
explicitly seek versions of lower complexity. Furthermore, these
systems are rule based and do not use learning approaches, the major
focus of this paper. In general, there has been very little
exploration of statistical machine learning techniques in these
fields, one of the few attempts being the recent work of Bridge \etal \cite{Bridge14} who use learning
to select between different heuristics for 1st order reasoning.
In contrast, our approach does not use hand-designed
heuristics, instead learning them automatically from the results of simpler
expressions. 

The attribute grammar, originally developed in 1968 by Knuth \cite{knuth1968semantics} in context of compiler
construction, has been successfully used as a tool for design and formal specification.
In our work, we apply attribute grammars to a search and optimization
problem. This has previously been explored in a range of domains: from
well-known algorithmic problems like knapsack packing
\cite{o2004solving}, through bioinformatics
\cite{waldispuhl2002approximate} to music \cite{desainte1994using}.
However, we are not aware of any previous work related to discovering
mathematical formulas using grammars, and learning in such framework. 
The closest work to ours can be
found in \cite{cheung1999attribute} which involves searching over the
space of algorithms and the grammar attributes also represent
computational complexity.

Classical techniques in natural language processing make extensive use
of grammars, for example to parse sentences and translate between
languages.  In this paper, we borrow techniques from NLP and apply
them to symbolic computation. In particular, we make use of an
$n$-gram model over mathematical operations, inspired by $n$-gram
language models. Recursive neural networks have also been recently
used in NLP, for example by Luong \etal \cite{luong2013better} and
Socher \etal \cite{socher2010learning,socher2013recursive}, as well as generic knowledge representation
Bottou \cite{bottou2014machine}. In particular, Socher \etal
\cite{socher2013recursive}, apply them to parse trees for sentiment
analysis. By contrast, we apply them to trees of symbolic expressions.
Our work also has similarities to Bowman \cite{bowman2013can} who shows that a recursive network can
learn simple logical predicates.

Our demonstration of continuous embeddings for symbolic expressions
has parallels with the embeddings used in NLP for words and sentence
structure, for example, Collobert \& Weston
\cite{Collobert08}, Mnih \& Hinton \cite{Mnih09}, Turian \etal \cite{Turian10}
and Mikolov \etal \cite{mikolov2013efficient}.

\section{Problem Statement}

{\noindent \bf Problem Definition:} We are given a symbolic {\em target expression}
$\mathbb{T}$ that combines a set of variables $\mathcal{V}$ to produce
an output $\mathbb{O}$, i.e. $\mathbb{O}=\mathbb{T}(\mathcal{V})$. We seek an alternate
expression $\mathbb{S}$, such that $\mathbb{S}(\mathcal{V})=\mathbb{T}(\mathcal{V})$, but has lower
computational complexity, i.e.~$\BigO{\mathbb{S}} < \BigO{\mathbb{T}}$. 

In this paper we consider the restricted setting where: (i)
$\mathbb{T}$ is a homogeneous polynomial of degree $k$\footnote{I.e.~It
  only contains terms of degree $k$. E.g. $ab + a^2 + ac$ is a
  homogeneous polynomial of degree 2, but $a^2 + b$ is not homogeneous
  ($b$ is of degree 1, but $a^2$ is of degree 2).}, (ii) $\mathcal{V}$
contains a single matrix or vector $A$ and (iii) $\mathbb{O}$ is a
scalar. While these assumptions may seem quite restrictive, they still
permit a rich family of expressions for our algorithm to
explore. For example, by combining multiple polynomial terms,
an efficient Taylor series approximation can be found for expressions
involving trigonometric or exponential operators. Regarding (ii), our framework can easily handle multiple
variables, e.g.~Figure \ref{fig:example_ab}, which shows expressions
using two matrices, $A$ and $B$. However, the rest of the paper
considers targets based on a single variable. In Section
\ref{sec:discussion}, we discuss these restrictions further.

{\noindent \bf Notation:} We adopt Matlab-style syntax for expressions.

\section{Attribute Grammar}
We first define an {\em attribute grammar} $\mathcal{G}$, which contains
a set of mathematical operations, each with an associated
complexity (the attribute). Since $\mathbb{T}$ contains exclusively polynomials, we
use the grammar rules listed in \tab{grammar}.  
\begin{table}
\scriptsize
\centering
\begin{tabular}{|l|l|l|l|l|}
\hline
Rule & Input & Output & Computation & Complexity\\ \hline \hline
Matrix-matrix multiply & $X \in \mathbb{R}^{n \times m}$ , $Y \in \mathbb{R}^{m \times p}$ &
$Z \in \mathbb{R}^{n \times p}$ & \texttt{Z = X * Y}  & $\BigO{nmp}$  \\ \hline
Matrix-element multiply & $X \in \mathbb{R}^{n \times m}$ , $Y \in \mathbb{R}^{n \times m}$ &
$Z \in \mathbb{R}^{n \times m}$ & \texttt{Z = X .* Y}  & $\BigO{nm}$  \\ \hline
Matrix-vector multiply & $X \in \mathbb{R}^{n \times m}$ , $Y \in \mathbb{R}^{m \times 1}$ &
$Z \in \mathbb{R}^{n \times n}$ & \texttt{Z = X * Y}  & $\BigO{nm}$  \\ \hline
Matrix transpose & $X \in \mathbb{R}^{n \times m}$ &
$Z \in \mathbb{R}^{m \times n}$ & \texttt{Z = X$^T$}  & $\BigO{nm}$  \\ \hline
Column sum  &  $X \in \mathbb{R}^{n \times m}$ &
$Z \in \mathbb{R}^{n \times 1}$ & \texttt{Z = sum(X,1)}  & $\BigO{nm}$  \\ \hline
Row sum  & $X \in \mathbb{R}^{n \times m}$ &
$Z \in \mathbb{R}^{1 \times m}$ & \texttt{Z = sum(X,2)}  & $\BigO{nm}$  \\ \hline
Column repeat  & $X \in \mathbb{R}^{n \times 1}$ &
$Z \in \mathbb{R}^{n \times m}$ & \texttt{Z = repmat(X,1,m)}  & $\BigO{nm}$  \\ \hline
Row repeat &  $X \in \mathbb{R}^{1 \times m}$ &
$Z \in \mathbb{R}^{n \times m}$ & \texttt{Z = repmat(X,n,1)}  & $\BigO{nm}$  \\ \hline
Element repeat &  $X \in \mathbb{R}^{1 \times 1}$ &
$Z \in \mathbb{R}^{n \times m}$ & \texttt{Z = repmat(X,n,m)}  & $\BigO{nm}$  \\ \hline
\end{tabular}
\vspace{-1.5mm}
\caption{The grammar $\mathcal{G}$ used in our experiments.}
\label{tab:grammar}
\vspace{-4mm}
\end{table}

Using these rules we can develop trees that combine rules to form
expressions involving $\mathcal{V}$, which for the purposes of this
paper is a single matrix $A$. Since we know $\mathbb{T}$ involves
expressions of degree $k$, each tree must use $A$ exactly $k$
times. Furthermore, since the output is a scalar, each tree must also
compute a scalar quantity. These two constraints limit the depth of
each tree. For some targets $\mathbb{T}$ whose complexity is only
$\BigO(n^3)$, we remove the matrix-matrix multiply rule, thus ensuring
that if any solution is found its complexity is at most $\BigO(n^2)$
(see \secc{results} for more details).  Examples of trees are shown in
\fig{example_ab}. The search strategy for determining which rules to
combine is addressed in Section \ref{sec:search_strategy}.

\vspace{-0.5mm}
\section{Representation of Symbolic Expressions}
\vspace{-1mm} 
We need an efficient way to check if the expression produced
by a given tree, or combination of trees (see \secc{linear}), matches $\mathbb{T}$. 
The conventional approach would be to perform this check symbolically,
but this is too slow for our purposes and is not amenable to
integration with learning methods. We therefore explore two alternate approaches.

\vspace{-1mm} 
\subsection{Numerical Representation}
\vspace{-1mm} 
\label{sec:numerical}
In this representation, each expression is represented by its evaluation of a randomly drawn set of $N$ 
points, where $N$ is large (typically $1000$). More precisely, for each
variable in $\mathcal{V}$, $N$ different copies are made, each populated
with randomly drawn elements. The target expression evaluates each of
these copies, producing a scalar value for each, so yielding a
vector $t$ of length $N$ which uniquely characterizes
$\mathbb{T}$. Formally, $t_n = \mathbb{T}(\mathcal{V}_n)$. We call
this numerical vector $t$ the {\em descriptor} of the symbolic expression $\mathbb{T}$.
The size of the descriptor $N$, must be sufficiently large to ensure
that different expressions are not mapped to the same
descriptor. Furthermore, when the descriptors are used in the
linear system of \eqn{linear} below, $N$ must also be greater than the
number of linear equations.
Any expression $\mathbb{S}$ formed by the grammar can be used to
evaluate each $\mathcal{V}_n$ to produce another $N$-length
descriptor vector $s$,
which can then be compared to $t$. If the two match, then
$\mathbb{S}(\mathcal{V})=\mathbb{T}(\mathcal{V})$.

In practice, using floating point values can result in numerical
issues that prevent $t$ and $s$ matching, even if the two expressions
are equivalent. We therefore use an integer-based descriptor in
the form of $\mathbb{Z}_p$\footnote{Integers modulo $p$}, 
where $p$ is a large prime number. This prevents both rounding issues
as well as numerical overflow.  

\vspace{-1mm} 
\subsection{Learned Representation}
\vspace{-1mm} 
\label{sec:learned}
We now consider how to learn a continuous representation for symbolic
expressions, that is learn a projection $\phi$ which maps expressions
$\mathbb{S}$ to $l$-dimensional vectors: $\phi(\mathbb{S}) \rightarrow \mathbb{R}^l$.  We use a recursive neural network (RNN) to do this,
in a similar fashion to Socher \etal \cite{socher2013recursive} for
natural language and Bowman \etal \cite{bowman2013can} for logical expressions.  This potentially allows many symbolic tasks to be
performed by machine learning techniques, in the same way that
the word-vectors (e.g.\cite{Collobert08} and \cite{mikolov2013efficient}) enable many NLP tasks to be posed a
learning problems.

We first create a dataset of symbolic expressions, spanning the space
of all valid expressions up to degree $k$. We then group them into
clusters of equivalent expressions (using the numerical representation
to check for equality), and give each cluster a discrete label $1
\ldots C$. For example, $A$, $(A^T)^T$ might have label 1, and $\sum_i
\sum_j A_{i, j}$, $\sum_j \sum_i A_{i, j}$ might have label 2 and so
on. For $k=6$, the dataset consists of $C=1687$ classes, examples of
which are show in \fig{expr}. Each class is split 80/20 into
train/test sets.

We then train a recursive neural network (RNN) to classify a grammar
tree into one of the $C$ clusters. Instead of
representing each grammar rule by its underlying arithmetic, we
parameterize it by a weight matrix or tensor (for operations with one
or two inputs, respectively) and use this to learn the {\em concept}
of each operation, as part of the network. A vector $a \in
\mathbb{R}^l$, where $l=30$\footnote{This was selected by
  cross-validation to control the capacity of the RNN, since it
  directly controls the number of parameters in the model.} is used to represent each input variable.
Working along the grammar tree, each operation in $\mathbb{S}$ evolves this
vector via matrix/tensor multiplications (preserving its length) until the entire
expression is parsed, resulting in a single vector $\phi(\mathbb{S})$ of length $l$, which is
passed to the classifier to determine the class of the expression, and
hence which other expressions it is equivalent to. 

\begin{figure}[h!]
\vspace{-5.5mm}
\begin{center}
\begin{subfigure}[b]{0.45\textwidth}
\includegraphics[width=\textwidth]{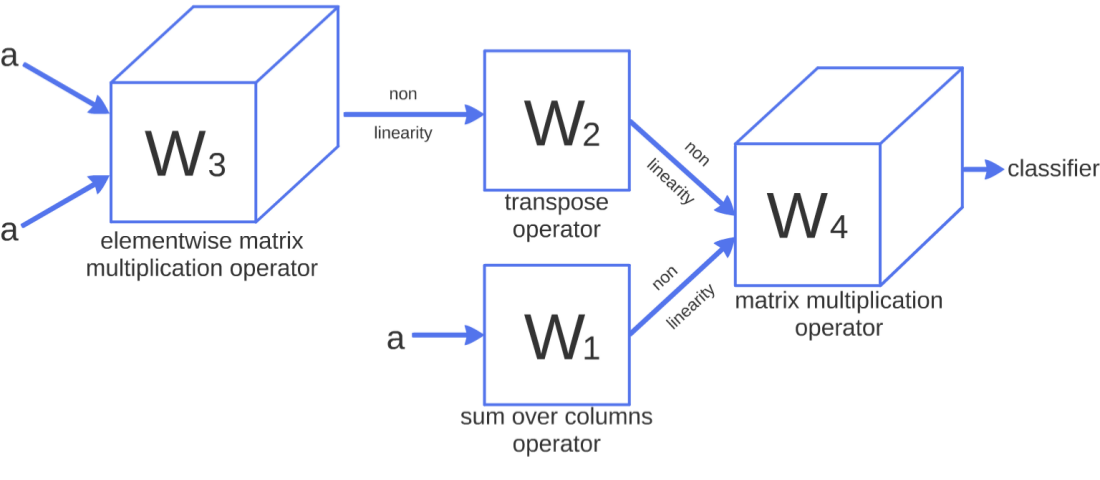}
\caption{$(A .* A)' * \text{sum}(A, 2)$}.
\end{subfigure}
\quad
\quad
\quad
\begin{subfigure}[b]{0.45\textwidth}
\includegraphics[width=\textwidth]{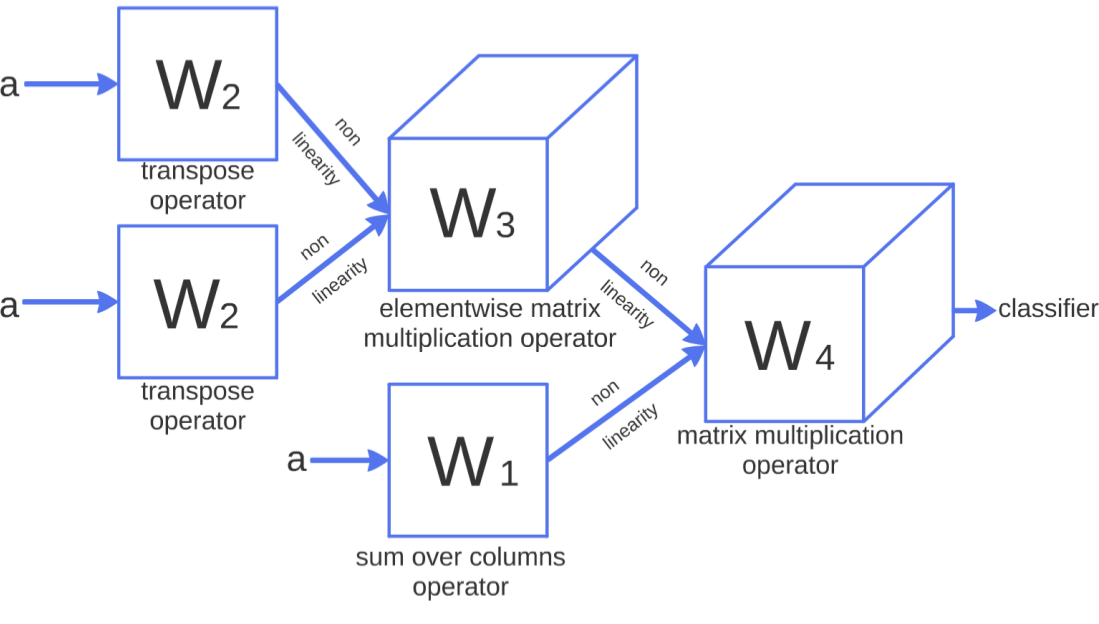}
\caption{ $(A' .* A') * \text{sum}(A, 2)$}.
\end{subfigure}
\end{center}
\vspace{-5mm}
\caption{Our RNN applied to two expressions. The matrix $A$
  is represented by a fixed random vector $a$ (of length $l=30$). Each
  operation in the expression applies a different matrix (for single
  input operations) or tensor (for dual inputs, e.g. matrix-element multiplication) to this vector. After
  each operation, a rectified-linear non-linearity is applied. The weight
  matrices/tensors for each operation are shared across different
  expressions. The final vector is passed to a softmax classifier (not
  shown) to predict which class they belong to. In this example, both
  expressions are equivalent, thus should be mapped to the same class.}
\label{fig:rnn}
\vspace{-2mm}
\end{figure}

\fig{rnn} shows this procedure for two different expressions. Consider
the first expression $\mathbb{S} =(A .* A)' * \text{sum}(A, 2)$. The first
operation here is $.*$, which is implemented in the RNN by taking the two (identical) vectors $a$
and applies a weight tensor $W_3$ (of size $l \times l \times
l$, so that the output is also size $l$), followed by a rectified-linear non-linearity. The output
of this stage is this $\max((W_3*a)*a,0)$. This vector is 
presented to the next operation, a matrix transpose, whose output is
thus $\max(W_2*\max((W_3*a)*a,0),0)$. Applying the remaining operations
produces a final output:
$\phi(\mathbb{S}) = \max((W_4*\max(W_2*\max((W_3*a)*a,0),0))*max(W_1*a,0))$. This is
presented to a $C$-way softmax classifier to predict the class of
the expression. The weights $W$ are trained using a
cross-entropy loss and backpropagation.

\begin{figure}[h!]
\vspace{-2mm}
\begin{center}
  \begin{subfigure}[b]{0.48\textwidth}
\miniscule
\begin{verbatim}
(((sum((sum((A * (A')), 1)), 2)) * ((A * (((sum((A'), 1)) * A)'))')) * A)
(sum(((sum((A * (A')), 2)) * ((sum((A'), 1)) * (A * ((A') * A)))), 1)) 
(((sum(A, 1)) * (((sum(A, 2)) * (sum(A, 1)))')) * (A * ((A') * A)))
((((sum((sum((A * (A')), 1)), 2)) * ((sum((A'), 1)) * (A * ((A') * A))))')')
((sum(A, 1)) * (((A') * (A * ((A') * ((sum(A, 2)) * (sum(A, 1))))))'))
((sum((sum((A * (A')), 1)), 2)) * ((sum((A'), 1)) * (A * ((A') * A))))
(((sum((sum((A * (A')), 1)), 2)) * ((sum((A'), 1)) * A)) * ((A') * A)) 
\end{verbatim}
\vspace{-2mm}
  \caption{Class A}
 \end{subfigure}
  \begin{subfigure}[b]{0.48\textwidth}
\miniscule
  \begin{verbatim}
 ((A') * ((sum(A, 2)) * ((sum((A'), 1)) * (A * (((sum((A'), 1)) * A)')))))
(sum(((A') * ((sum(A, 2)) * ((sum((A'), 1)) * (A * ((A') * A))))), 2)) 
((((sum(A, 2)) * ((sum((A'), 1)) * A))') * (A * (((sum((A'), 1)) * A)')))
(((sum((A'), 1)) * (A * ((A') * ((sum(A, 2)) * ((sum((A'), 1)) * A)))))')
((((sum((A'), 1)) * A)') * ((sum((A'), 1)) * (A * (((sum((A'), 1)) * A)'))))
(((A * ((A') * ((sum(A, 2)) * ((sum((A'), 1)) * A))))') * (sum(A, 2)))
(((A') * ((sum(A, 2)) * ((sum((A'), 1)) * A))) * (sum(((A') * A), 2)))
\end{verbatim}
\vspace{-2mm}
  \caption{Class B}
  \end{subfigure}
\end{center}
\vspace{-3mm}
\caption{Samples from two classes of degree $k=6$ in our dataset of expressions, used
  to learn a continuous representation of symbolic
  expressions via an RNN. Each line represents a different expression, but those
  in the same class are equivalent to one another. }
\label{fig:expr}
\end{figure}

When training the RNN, there are several important details that are
crucial to obtaining high classification accuracy:
\begin{compactitem}
\item The weights should be initialized to the identity, plus a small
  amount of Gaussian noise added to all elements. The identity allows
information to flow the full length of the network, up to the classifier regardless of
its depth \cite{saxe2013exact}. Without this, the RNN overfits badly,
producing test accuracies of $\sim 1\%$.
\item Rectified linear units work much better in this setting than tanh
  activation functions.
\item We learn using a curriculum \cite{Bengio09, zaremba2014learning}, starting with the simplest
  expressions of low degree and slowly increasing $k$.  
\item The weight matrix in the softmax classifier has much larger
  ($\times100$) learning rate than the rest of the layers. This
  encourages the representation to stay still even when targets are
  replaced, for example, as we move to harder examples.
\item As well as updating the weights of the RNN, we also update the
  initial value of $a$ (i.e~we backpropagate to the input also).
\end{compactitem}
When the RNN-based representation is employed for identity discovery (see
\secc{rnn}), the vector $\phi(\mathbb{S})$ is used directly (i.e.~ the
$C$-way softmax used in training is removed from the network).

\section{Linear Combinations of Trees}
\label{sec:linear}
For simple targets, an expression that matches the target may be contained within a single
grammar tree. But more complex expressions typically require
a linear combination of expressions from different trees. 

To handle this, we can use the integer-based
descriptors for each tree in a linear system 
and solve for a match to the target descriptor (if one exists).   
Given a set of $M$ trees, each with its own integer descriptor vector
$f$, we form an $M$ by $N$ linear system of equations and solve
it:
\begin{equation*}
Fw = t \text{ mod } \mathbb{Z}_p
\label{eqn:linear}
\end{equation*}
where $F = [f_1,\ldots, f_M]$ holds the tree representations, $w$ is
the weighting on each of the trees and $t$ is the target
representation. The system is solved using Gaussian elimination,
where addition and multiplication is performed
modulo $p$. 
The number of solutions can vary: (a) there can be {\bf no solution},
which means that no linear combination of
the current set of trees can match the target expression. If all
possible trees have been enumerated, then this implies the target
expression is outside the scope of the grammar. (b) There can be {\bf
  one or more solutions}, meaning that some combination of the
current set of trees yields a match to the target expression.

\vspace{-2mm}
\section{Search Strategy}
\vspace{-1mm}
\label{sec:search_strategy}
So far, we have proposed a grammar which defines the computations
that are permitted (like a programming language grammar), but it gives
no guidance as to how explore the space of possible expressions. Neither
do the representations we introduced help -- they simply allow us to
determine if an expression matches or not. We now describe how to
efficiently explore the space by learning which paths are likely to
yield a match.

Our framework uses two components: a {\bf scheduler}, and a {\bf
  strategy}. The scheduler is fixed, and traverses space of
expressions according to recommendations given by the selected
strategy (e.g.~``Random'' or ``$n$-gram'' or ``RNN''). The
strategy assesses which of the possible grammar rules is likely to
lead to a solution, given the current expression. Starting with 
the variables $\mathcal{V}$ (in our case a single element $A$, or more
generally, the elements $A$, $B$ etc.), at each step the scheduler receives 
scores for each rule from the strategy and picks the one with the highest
score. This continues until the expression reaches degree $k$ and the tree is
complete. We then run the linear solver to see if a linear combination
of the existing set of trees matches the target. If not, the scheduler
starts again with a new tree, initialized with the set of variables $\mathcal{V}$.
The $n$-gram and RNN strategies are learned in an incremental fashion, starting with simple
target expressions (i.e.~those of low degree $k$, such as $\sum_{ij}
AA^T$). Once solutions to these are found, they become training
examples used to improve the strategy, needed for tackling harder targets (e.g.~$\sum_{ij}
AA^TA$).

\vspace{-1mm}
\subsection{Random Strategy}
\vspace{-1mm} The random strategy involves no learning, thus assigns
equal scores to all valid grammar rules, hence the scheduler randomly
picks which expression to try at each step. For simple targets, this
strategy may succeed as the scheduler may stumble upon a match to the
target within a reasonable time-frame. But for complex target
expressions of high degree $k$, the search space is huge and the
approach fails.

\vspace{-1mm}
\subsection{$n$-gram}
\vspace{-1mm}
In this strategy, we simply count how often subtrees of depth $n$
occur in solutions to previously solved targets. As the number of
different subtrees of depth $n$ is large, the counts become very
sparse as $n$ grows. Due to this, we use a weighted linear combination
of the score from all depths up to $n$. We found an effective
weighting to be $10^k$, where $k$ is the depth of the
tree. 

\vspace{-1mm}
\subsection{Recursive Neural Network}
\vspace{-1mm}
\label{sec:rnn}
Section \ref{sec:learned} showed how to use an RNN to learn a continuous
representation of grammar trees. Recall that the RNN $\phi$
maps expressions to continuous vectors: $\phi(\mathbb{S}) \rightarrow \mathbb{R}^l$.
To build a search strategy from this, we train a 
softmax layer on top of the RNN to predict which rule should be
applied to the current expression (or expressions, since some rules
have two inputs), so that we match the target.

Formally, we have two current branches $b_1$ and $b_2$ (each
corresponding to an expression) and wish to predict
the root operation $r$ that joins them (e.g.~$.*$) from among the
valid grammar rules ($|r|$ in total). We first use the previously trained RNN to compute 
$\phi(b_1)$ and $\phi(b_2)$. These are then presented to a $|r|$-way
softmax layer (whose weight matrix $U$ is of size $2l \times |r|$). 
If only one branch exists, then $b_2$ is set to a fixed random
vector. The training data for $U$ comes from trees that
give efficient solutions to targets of lower degree $k$ (i.e.~simpler
targets). Training of the softmax layer is performed by stochastic
gradient descent. We use dropout \cite{hinton2012improving} as the
network has a tendency to overfit and repeat exactly the same
expressions for the next value of $k$. Thus, instead of training on
exactly $\phi(b_1)$ and $\phi(b_2)$, we drop activations as we
propagate toward the top of the tree (the same fraction for each
depth), which encourages the RNN to capture more local structures.  At
test time, the probabilities from the softmax become the scores used
by the scheduler.

\vspace{-1mm}
\section{Experiments}
\vspace{-2mm} 
We first show results relating to the learned representation for
symbolic expressions (\secc{learned}). Then we demonstrate our
framework discovering efficient identities. For brevity, the
identities discovered are listed in the supplementary material \cite{arxiv}. 

\vspace{-1mm}
\subsection{Expression Classification using Learned Representation}
\vspace{-1mm}
\tab{classify} shows the accuracy of the RNN model on expressions of
varying degree, ranging from $k=3$ to $k=6$. The difficulty of the
task can be appreciated by looking at the examples in \fig{expr}. The
low error rate of $\leq 5\%,$ despite the use of a simple softmax
classifier, demonstrates the effectiveness of our learned representation. 
\begin{table}[h!]
  \centering
 \vspace{-2mm}
  \scriptsize
  \begin{tabular}{lcccc}
    \hline
                           & Degree $k=3$      &  Degree $k=4$          & Degree $k=5$    & Degree $k=6$  \\
    \hline
    Test accuracy          & $100\% \pm 0\%$ &   $96.9\% \pm 1.5\%$ &  $94.7\% \pm 1.0\%$     &    $95.3\% \pm 0.7\%$ \\
    Number of classes      &           12    &            125       &      970        &     1687      \\
    Number of expressions &  126 &  1520 & 13038 & 24210 \\
 \end{tabular}
  \caption{Accuracy of predictions using our learned
    symbolic representation (averaged over $10$ different initializations). As the degree increases tasks becomes more challenging, 
  because number of classes grows, and computation trees become
  deeper. However our dataset grows larger too (training uses 80\% of
  examples).}
\label{tab:classify}
\end{table}

\vspace{-4mm}
\subsection{Efficient Identity Discovery}
\label{sec:results}
\begin{figure}[h!]
\vspace{-3mm}
\begin{center}
\includegraphics[width=0.24\linewidth]{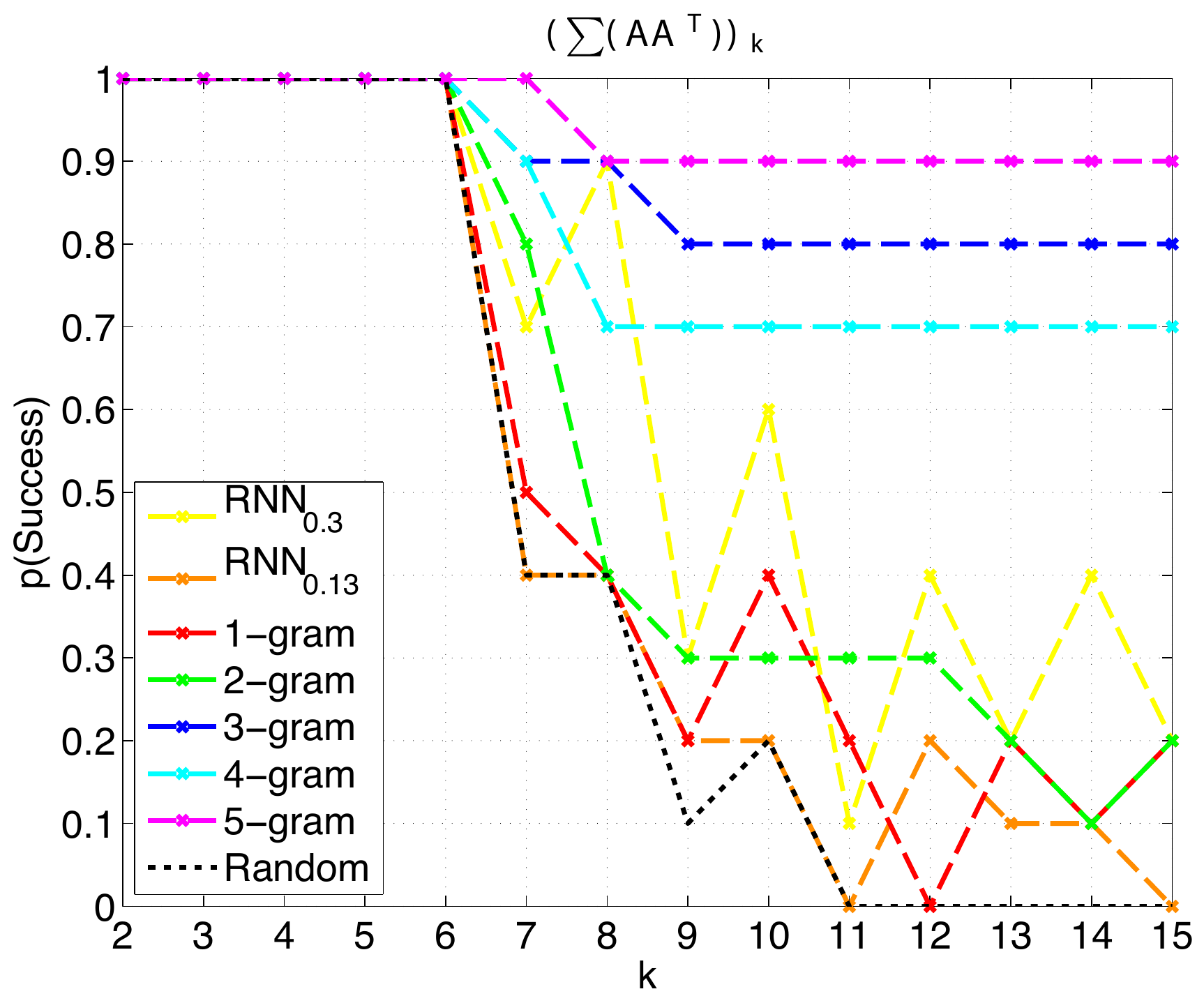}
\includegraphics[width=0.24\linewidth]{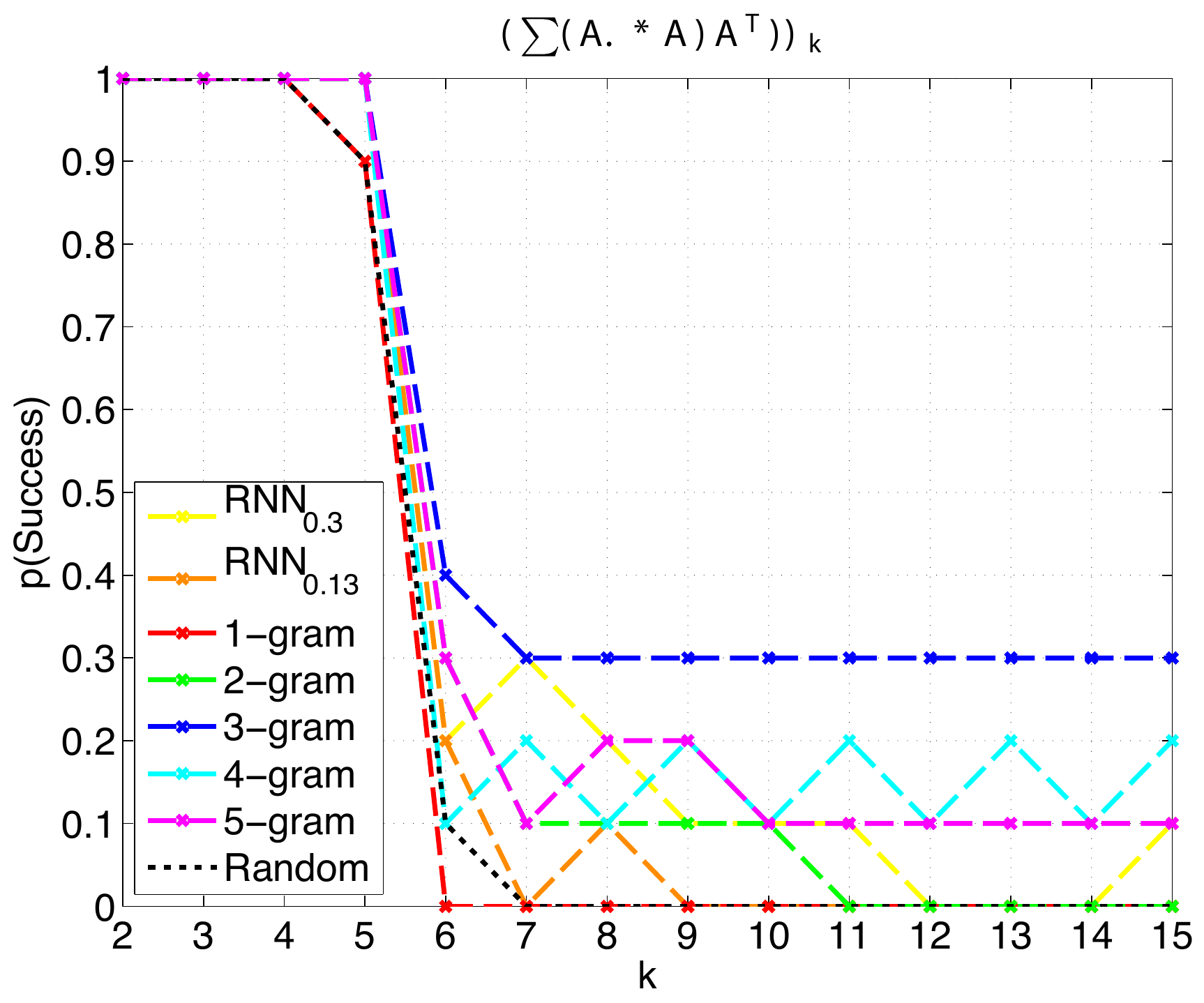}
\includegraphics[width=0.24\linewidth]{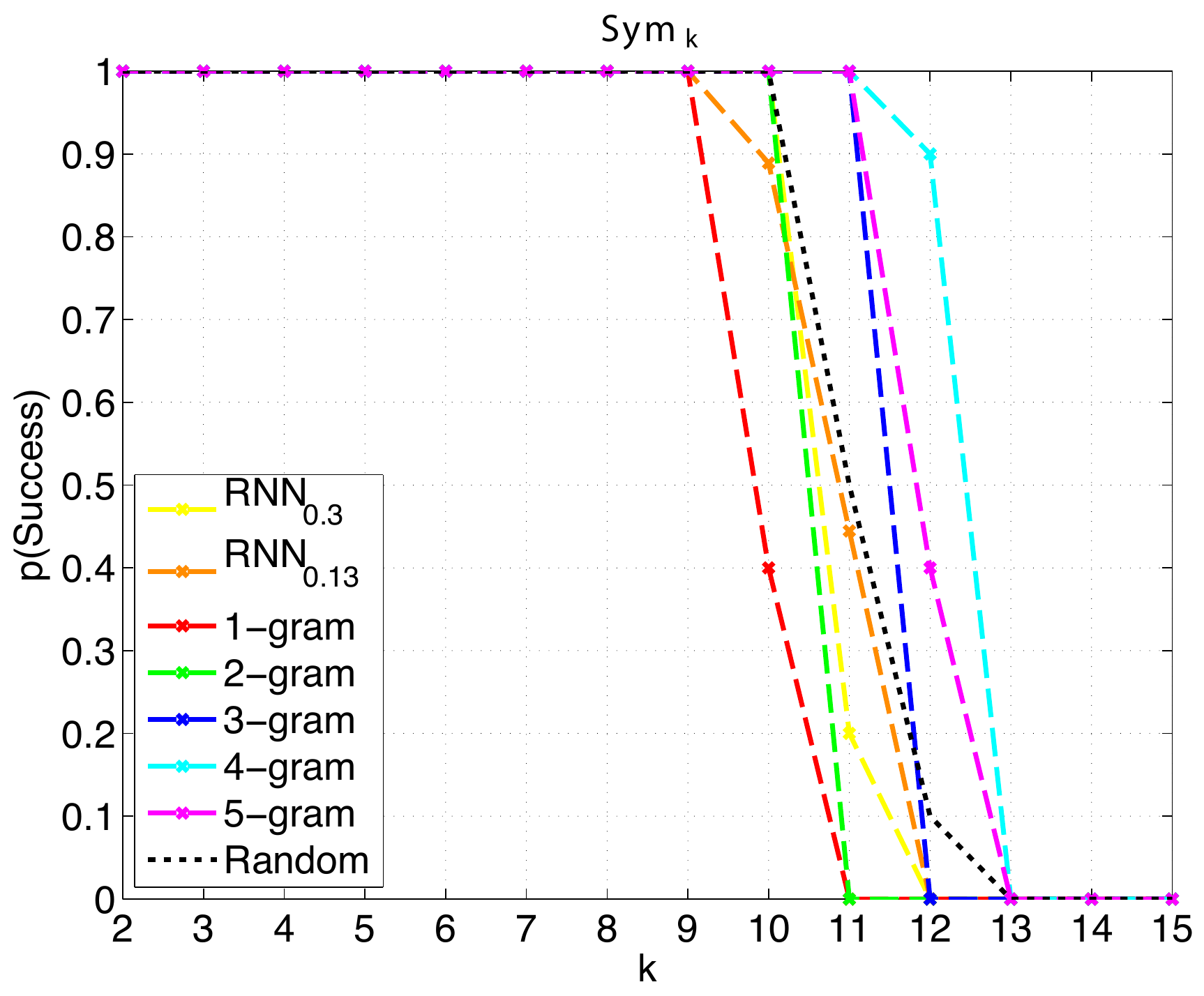}
\includegraphics[width=0.24\linewidth]{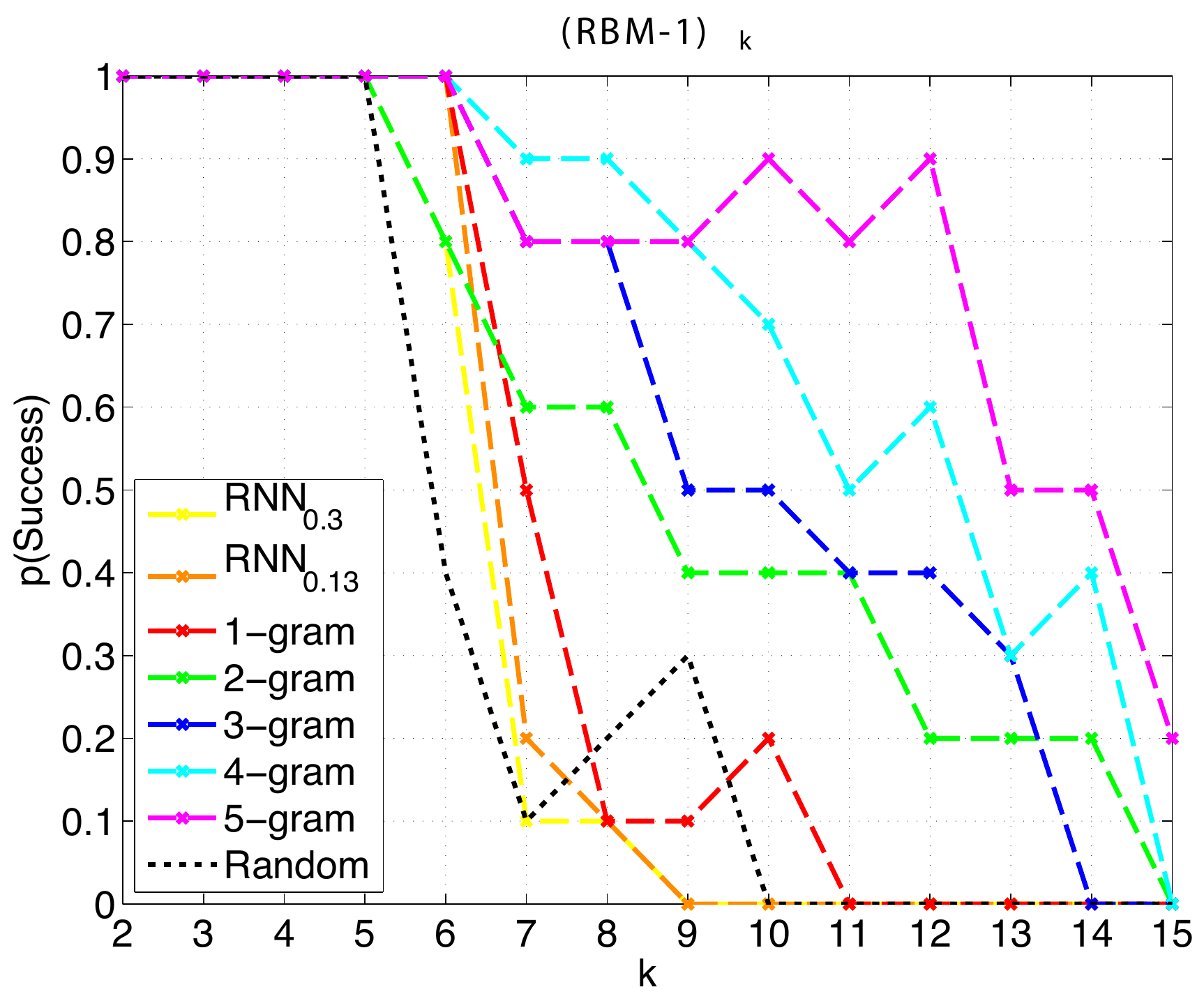}
\end{center}
\vspace*{-0.3cm}
\caption{Evaluation on four different families of expressions. As the
  degree $k$ increases, we see that the random strategy consistently fails but the
  learning approaches can still find solutions (i.e.~p(Success) is
  non-zero). Best viewed in electronic form.}
\label{fig:induction}
\vspace{-3mm}
\end{figure}

In our experiments we consider 5 different families of expressions,
chosen to fall within the scope of our grammar rules: 
\vspace{-2mm}
\begin{enumerate}
  \item {$\mathbf{(\sum AA^T)_k}$:} $A$ is an $\mathbb{R}^{n \times
      n}$ matrix. The $k$-th term is $\sum_{i,j}
    (AA^T)^{\lfloor k/2\rfloor}$ for even $k$ \\ and $\sum_{i,j}
    (AA^T)^{\lfloor k/2\rfloor}A$ , for odd $k$.  E.g. for $k=2: \sum_{i,j} AA^T$;
    for $k=3: \sum_{i,j} AA^TA$; for $k=4: \sum_{i, j} AA^TAA^T$
    etc. Naive evaluation is $\BigO{kn^3}$.
\vspace{-1mm}
  \item {$(\mathbf{\sum (A.*A)A^T})_k$:} $A$ is an $\mathbb{R}^{n \times
      n}$ matrix and let $B = A.*A$. The $k$-th term is $\sum_{i,j}
    (BA^T)^{\lfloor k/2\rfloor}$ for even $k$ and $\sum_{i,j}
    (BA^TB)^{\lfloor k/2\rfloor}$ , for odd $k$.  E.g. for $k=2: \sum_{i,j} (A.*A)A^T$;
    for $k=3: \sum_{i,j} (A.*A)A^T(A.*A)$; for $k=4: \sum_{i, j}(A.*A)A^T(A.*A)A^T$
    etc. Naive evaluation is $\BigO{kn^3}$.
\vspace{-1mm}
 \item {\bf Sym$_k$}: Elementary symmetric polynomials. $A$ is a
   vector in $\mathbb{R}^{n \times 1}$.  For $k=1: \sum_{i}A_i$, for $k=2:
   \sum_{i<j}A_iA_j$, for $k=3: \sum_{i<j<k}A_iA_jA_k$, etc. Naive
   evaluation is $\BigO{n^k}$. 
\vspace{-1mm}
  \item {\bf (RBM-1)$_k$}: $A$ is a vector in $\mathbb{R}^{n \times 1}$. $v$ is
    a binary $n$-vector. The $k$-th term is: $\sum_{v \in \{0, 1\}^n}
    (v^T A)^k$. Naive evaluation is $\BigO{2^n}$.
\vspace{-1mm}
  \item {\bf (RBM-2)$_k$}: Taylor series terms for the partition
    function of an RBM. $A$ is a matrix in $\mathbb{R}^{n \times
    n}$. $v$ and $h$ are a binary $n$-vectors. The $k$-th term is $\sum_{v \in \{0,1\}^n,h \in \{0, 1\}^n} (v^TAh)^k$. Naive evaluation is $\BigO{2^{2n}}$.
\end{enumerate}
\vspace{-1mm}
Note that (i) for all families, the expressions yield a scalar output;
(ii) the families are ordered in rough order of ``difficulty''; (iii)
we are not aware of any previous exploration of these expressions,
except for {\bf Sym$_k$}, which is well studied \cite{stanley2011enumerative}. 
For the
$\mathbf{(\sum AA^T)_k}$ and $(\mathbf{\sum (A.*A)A^T})_k$ families we
remove the matrix-multiply rule from the grammar, thus ensuring that if any solution is
found it will be efficient since the remaining rules are at most
$\BigO{kn^2}$, rather than $\BigO{kn^3}$. The other families use the full
grammar, given in \tab{grammar}. However, the limited set of rules
means that if any solution is found, it can at most be $\BigO{n^3}$,
rather than exponential in $n$, as the naive evaluations would be. 
For each family, we apply our framework, using the three different
search strategies introduced in \secc{search_strategy}. 
For each run we impose a
fixed cut-off time of 10 minutes\footnote{Running on a 3Ghz 16-core
  Intel Xeon. Changing the cut-off has little effect on the plots,
  since the search space grows exponentially fast.} beyond which we
terminate the search. At each value of $k$, we repeat the experiments
10 times with different random initializations and count the number of
runs that find an efficient solution. Any non-zero count is deemed a
success, since each identity only needs to be discovered once. However,
in \fig{induction}, we show the fraction of successful runs, which
gives a sense of how quickly the identity was found. 

We start with $k=2$ and increase up to $k=15$, using the solutions
from previous values of $k$ as training data for the current
degree. The search space quickly grows with $k$, as shown in
\tab{num_exp}. \fig{induction} shows results for four of the
families. We use $n$-grams for $n=1 \ldots 5$, as well as the RNN with
two different dropout rates (0.125 and 0.3). The learning approaches
generally do much better than the random strategy for large values of
$k$, with the $3$-gram, $4$-gram and $5$-gram models outperforming the
RNN. 

For the first two families, the $3$-gram model reliably finds
solutions. These solutions involve repetition of a local patterns
(e.g.~Example 2), which can easily be captured with $n$-gram
models. However, patterns that don't have a simple repetitive
structure are much more difficult to generalize. The {\bf (RBM-2)$_k$}
family is the most challenging, involving a double exponential sum,
and the solutions have highly complex trees (see supplementary
material \cite{arxiv}). In this case, none of our approaches performed better than
the random strategy and no solutions were discovered for
$k>5$. However, the $k=5$ solution was found by the RNN consistently
faster than the random strategy ($100 \pm 12$ vs $438 \pm 77$ secs).

\begin{table}[h!]
\vspace{-2mm}
  \centering
  \scriptsize
  \begin{tabular}{lcccccc}
    \hline
                         & $k=2$   & $k=3$      &  $k=4$          & $k=5$    & $k=6$ & $k=7$ and higher \\
    \hline
    \# Terms $\leq \BigO{n^2}$ &  39 & 171 & 687 & 2628 & 9785 & \multirow{2}{*}{Out of memory}  \\
    \# Terms $\leq \BigO{n^3}$ &  41 & 187 & 790 & 3197 & 10k$+$ & \\
 \end{tabular}
\vspace{-1mm}
  \caption{The number of possible expressions for different degrees $k$.}
\label{tab:num_exp}
\end{table}

\vspace{-3mm}
\section{Discussion}
\vspace{-2mm}
\label{sec:discussion}
We have introduced a framework based on a grammar of symbolic
operations for discovering mathematical identities. Through the novel
application of learning methods, we have shown how the exploration of
the search space can be learned from previously successful solutions
to simpler expressions. This allows us to discover complex expressions
that random or brute-force strategies cannot find (the identities are
given in the supplementary material \cite{arxiv}). 

Some of the families considered in this paper are close to
expressions often encountered in machine learning. For example,
dropout involves an exponential sum over binary masks, which is
related to the {\bf RBM-1} family. Also, the partition function of an
RBM can be approximated by the {\bf RBM-2} family. Hence the identities
we have discovered could potentially be used to give a closed-form
version of dropout, or compute the RBM partition function efficiently
(i.e.~in polynomial time). 
Additionally, the automatic nature of our system naturally lends itself to
integration with compilers, or other optimization tools, where it could replace computations with
efficient versions thereof.

Our framework could potentially be applied to more general settings,
to discover novel formulae in broader areas of mathematics. To
realize this, additional grammar rules, e.g.~involving
recursion or trigonometric functions would be needed. However, this
would require a more complex scheduler to determine when to terminate
a given grammar tree. 
Also, it is surprising that a recursive neural network can generate an
effective continuous representation for symbolic expressions. This
could have broad applicability in allowing machine learning tools to
be applied to symbolic computation.

The problem addressed in this paper involves discrete search within a
combinatorially large space -- a core problem with AI. Our successful
use of machine learning to guide the search gives hope that similar
techniques might be effective in other AI tasks where combinatorial
explosions are encountered.

\vspace{-2mm}
\section*{Acknowledgements}
\vspace{-2mm} 
The authors would like to thank Facebook and Microsoft Research for their support.  

\small
\bibliography{bibliography}
\bibliographystyle{ieee}



\section*{Supplementary material}

## 9 Supplementary material

We present the efficient expressions discovered by our system, using Matlab-style syntax, and we visualize computation trees. Each example contains: (i) code that computes the original target formulas; (ii) the formulae derived by our system and (iii) code that verifies the correctness of the expression. The size of matrices $n$, $m$ can be chosen arbitrary.

Code for generating the expressions can be downloaded from https://github.com/kkurach/math_learning. The source files for this paper are available at https://github.com/kkurach/math_learning/paper/.

### 9.1 $(\sum AA^T)_k$

**k = 1**

```
n = 100;
m = 200;
A = randn(n, m);
original = sum(sum(A, 1), 2);

optimized = 1 * (sum(sum(A, 2), 1));
normalization = sum(abs(original(:)));
assert(sum(abs(original(:) - optimized(:))) / normalization < 1e-10);
```

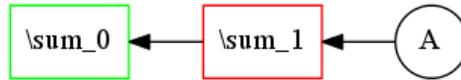

**k = 2**

```
n = 100;
m = 200;
A = randn(n, m);
original = sum(sum((A * A'), 1), 2);

optimized = 1 * (sum((sum(A, 1) * A'), 2));
normalization = sum(abs(original(:)));
assert(sum(abs(original(:) - optimized(:))) / normalization < 1e-10);
```

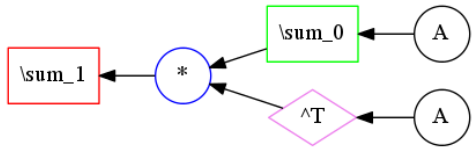

**k = 3**

```
n = 100;
m = 200;
A = randn(n, m);
original = sum(sum(((A * A') * A), 1), 2);

optimized = 1 * (sum((A * (sum(A, 2)' * A)'), 1));
normalization = sum(abs(original(:)));
assert(sum(abs(original(:) - optimized(:))) / normalization < 1e-10);
```

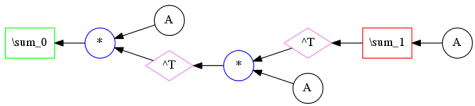

**k = 4**

```
n = 100;
m = 200;
A = randn(n, m);
original = sum(sum((((A * A') * A) * A'), 1), 2);

optimized = 1 * (((sum(A, 1) * A') * (sum(A, 1) * A')'));
normalization = sum(abs(original(:)));
assert(sum(abs(original(:) - optimized(:))) / normalization < 1e-10);
```

1

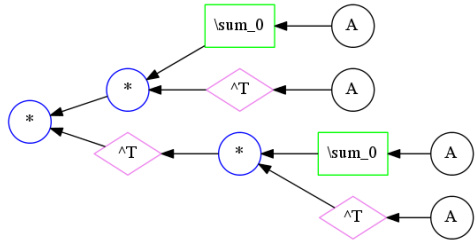

**k = 5**

```
n = 100;
m = 200;
A = randn(n, m);
original = sum(sum(((((A * A') * A) * A') * A), 1), 2);

optimized = 1 * (sum((A * ((A * (sum(A, 2)' * A)')' * A)'), 1));
normalization = sum(abs(original(:)));
assert(sum(abs(original(:) - optimized(:))) / normalization < 1e-10);
```

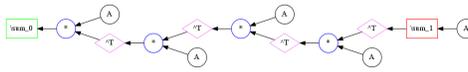

**k = 6**

```
n = 100;
m = 200;
A = randn(n, m);
original = sum(sum((((((A * A') * A) * A') * A) * A'), 1), 2);

optimized = 1 * (sum((A * ((A * ((sum(A, 1) * A') * A)')' * A)'), 1));
normalization = sum(abs(original(:)));
assert(sum(abs(original(:) - optimized(:))) / normalization < 1e-10);
```

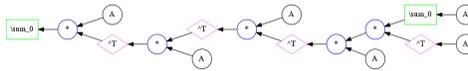

**k = 7**

```
n = 100;
m = 200;
A = randn(n, m);
original = sum(sum((((((A * A') * A) * A') * A) * A') * A), 1), 2);

optimized = 1 * (sum((A * ((A * ((A * (sum(A, 2)' * A)')' * A)')' * A)'), 1));
normalization = sum(abs(original(:)));
assert(sum(abs(original(:) - optimized(:))) / normalization < 1e-10);
```

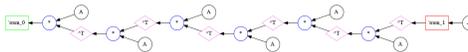

**k = 8**

```
n = 100;
m = 200;
A = randn(n, m);
original = sum(sum(((((((A * A') * A) * A') * A) * A') * A) * A'), 1), 2);

optimized = 1 * (sum((A * ((A * ((A * ((sum(A, 1) * A') * A)')' * A)')' * A)'), 1));
normalization = sum(abs(original(:)));
assert(sum(abs(original(:) - optimized(:))) / normalization < 1e-10);
```

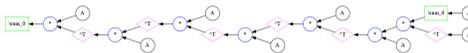

**k = 9**

```
n = 100;
m = 200;
A = randn(n, m);
original = sum(sum(((((((((A * A') * A) * A') * A) * A') * A) * A') * A), 1), 2);

optimized = 1 * (sum((A * ((A * ((A * ((A * (sum(A, 2)' * A)')' * A)')' * A)')' * A)'), 1));
normalization = sum(abs(original(:)));
assert(sum(abs(original(:) - optimized(:))) / normalization < 1e-10);
```

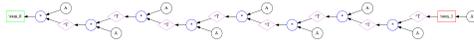

**k = 10**

```
n = 100;
m = 200;
A = randn(n, m);
original = sum(sum((((((((((A * A') * A) * A') * A) * A') * A) * A') * A) * A'), 1), 2);

optimized = 1 * (sum((A * ((A * ((A * ((A * ((sum(A, 1) * A') * A)')' * A)')' * A)')' * A)'), 1));
normalization = sum(abs(original(:)));
assert(sum(abs(original(:) - optimized(:))) / normalization < 1e-10);
```

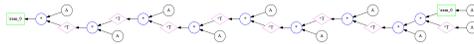

**k = 11**

```
n = 100;
m = 200;
A = randn(n, m);
original = sum(sum(((((((((((A * A') * A) * A') * A) * A') * A) * A') * A) * A') * A), 1), 2);

optimized = 1 * (sum((A * ((A * ((A * ((A * ((A * (sum(A, 2)' * A)')' * A)')' * A)')' * A)')' * A)'), 1));
normalization = sum(abs(original(:)));
assert(sum(abs(original(:) - optimized(:))) / normalization < 1e-10);
```

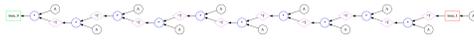

**k = 12**

```
n = 100;
m = 200;
A = randn(n, m);
original = sum(sum((((((((((((A * A') * A) * A') * A) * A') * A) * A') * A) * A') * A) * A'), 1), 2);

optimized = 1 * (sum((A * ((A * ((A * ((A * ((A * ((sum(A, 1) * A') * A)')' * A)')' * A)')' * A)')' * A)'), 1)
    );
normalization = sum(abs(original(:)));
assert(sum(abs(original(:) - optimized(:))) / normalization < 1e-10);
```

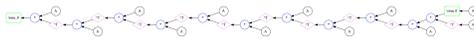

**k = 13**

```
n = 100;
m = 200;
A = randn(n, m);
original = sum(sum(((((((((((((A * A') * A) * A') * A) * A') * A) * A') * A) * A') * A) * A') * A), 1), 2);

optimized = 1 * (sum((A * ((A * ((A * ((A * ((A * ((A * (sum(A, 2)' * A)')' * A)')' * A)')' * A)')' * A)')' *
    A)'), 1));
normalization = sum(abs(original(:)));
assert(sum(abs(original(:) - optimized(:))) / normalization < 1e-10);
```

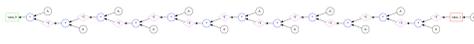

**k = 14**

3

```
n = 100;
m = 200;
A = randn(n, m);
original = sum(sum(((((((((((((A * A') * A) * A') * A) * A') * A) * A') * A) * A') * A) * A') * A) * A'), 1),
    ↪ 2);

optimized = 1 * (sum((A * ((A * ((A * ((A * ((A * ((A * ((sum(A, 1) * A') * A)')' * A)')' * A)')' * A)
    ↪ ')' * A)'), 1));
normalization = sum(abs(original(:)));
assert(sum(abs(original(:) - optimized(:))) / normalization < 1e-10);
```

**k = 15**

```
n = 100;
m = 200;
A = randn(n, m);
original = sum(sum(((((((((((((((A * A') * A) * A') * A) * A') * A) * A') * A) * A') * A) * A') * A) * A') * A
    ↪ ), 1), 2);

optimized = 1 * (sum((A * ((A * ((A * ((A * ((A * ((A * ((A * (sum(A, 2)' * A)')' * A)')' * A)')' * A)')' * A)
    ↪ ')' * A)')' * A)'), 1));
normalization = sum(abs(original(:)));
assert(sum(abs(original(:) - optimized(:))) / normalization < 1e-10);
```

## 9.2 $\left(\sum AB\right)_k$

**k = 1**

```
n = 100;
m = 200;
A = randn(n, m);
B = randn(m, n);
original = sum(sum(A, 1), 2);

optimized = 1 * (sum(sum(A, 1), 2));
normalization = sum(abs(original(:)));
assert(sum(abs(original(:) - optimized(:))) / normalization < 1e-10);
```

**k = 2**

```
n = 100;
m = 200;
A = randn(n, m);
B = randn(m, n);
original = sum(sum((A * B), 1), 2);

optimized = 1 * ((sum(A, 1) * sum(B, 2)));
normalization = sum(abs(original(:)));
assert(sum(abs(original(:) - optimized(:))) / normalization < 1e-10);
```

**k = 3**

```
n = 100;
m = 200;
A = randn(n, m);
B = randn(m, n);
original = sum(sum(((A * B) * A), 1), 2);

optimized = 1 * (sum(((sum(A, 1) * B) * A), 2));
normalization = sum(abs(original(:)));
assert(sum(abs(original(:) - optimized(:))) / normalization < 1e-10);
```

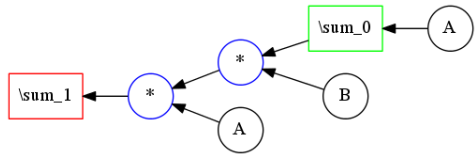

**k = 4**

```
n = 100;
m = 200;
A = randn(n, m);
B = randn(m, n);
original = sum(sum((((A * B) * A) * B), 1), 2);

optimized = 1 * (sum((((sum(A, 1) * B) * A) * B), 2));
normalization = sum(abs(original(:)));
assert(sum(abs(original(:) - optimized(:))) / normalization < 1e-10);
```

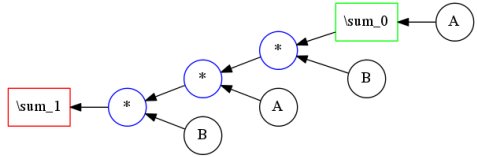

**k = 5**

```
n = 100;
m = 200;
A = randn(n, m);
B = randn(m, n);
original = sum(sum(((((A * B) * A) * B) * A), 1), 2);

optimized = 1 * (sum(((((sum(A, 1) * B) * A) * B) * A), 2));
normalization = sum(abs(original(:)));
assert(sum(abs(original(:) - optimized(:))) / normalization < 1e-10);
```

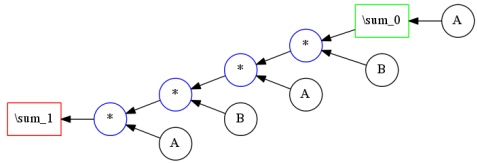

**k = 6**

```
n = 100;
m = 200;
A = randn(n, m);
B = randn(m, n);
original = sum(sum((((((A * B) * A) * B) * A) * B), 1), 2);

optimized = 1 * (sum((((((sum(A, 1) * B) * A) * B) * A) * B), 2));
normalization = sum(abs(original(:)));
assert(sum(abs(original(:) - optimized(:))) / normalization < 1e-10);
```

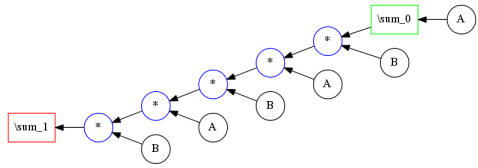

### k = 7

```
n = 100;
m = 200;
A = randn(n, m);
B = randn(m, n);
original = sum(sum((((((A * B) * A) * B) * A) * B) * A), 1), 2);

optimized = 1 * (sum(((((((sum(A, 1) * B) * A) * B) * A) * B) * A), 2));
normalization = sum(abs(original(:)));
assert(sum(abs(original(:) - optimized(:))) / normalization < 1e-10);
```

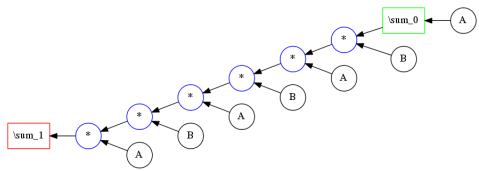

### k = 8

```
n = 100;
m = 200;
A = randn(n, m);
B = randn(m, n);
original = sum(sum(((((((A * B) * A) * B) * A) * B) * A) * B), 1), 2);

optimized = 1 * (sum((((((((sum(A, 1) * B) * A) * B) * A) * B) * A) * B), 2));
normalization = sum(abs(original(:)));
assert(sum(abs(original(:) - optimized(:))) / normalization < 1e-10);
```

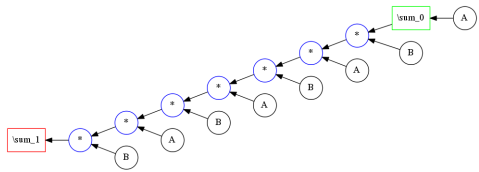

### k = 9

```
n = 100;
m = 200;
A = randn(n, m);
B = randn(m, n);
original = sum(sum((((((((A * B) * A) * B) * A) * B) * A) * B) * A), 1), 2);

optimized = 1 * (sum(((((((((sum(A, 1) * B) * A) * B) * A) * B) * A) * B) * A), 2));
normalization = sum(abs(original(:)));
assert(sum(abs(original(:) - optimized(:))) / normalization < 1e-10);
```

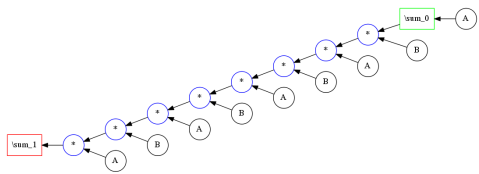

### k = 10

```
n = 100;
m = 200;
A = randn(n, m);
```

```
B = randn(m, n);
original = sum(sum(((((((((A * B) * A) * B) * A) * B) * A) * B) * A) * B), 1), 2);

optimized = 1 * (sum(((((((((sum(A, 1) * B) * A) * B) * A) * B) * A) * B) * A) * B), 2));
normalization = sum(abs(original(:)));
assert(sum(abs(original(:) - optimized(:))) / normalization < 1e-10);
```

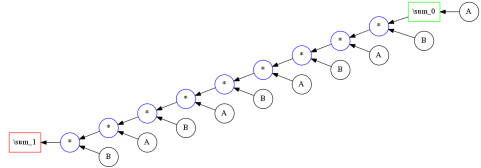

### k = 11

```
n = 100;
m = 200;
A = randn(n, m);
B = randn(m, n);
original = sum(sum((((((((((A * B) * A) * B) * A) * B) * A) * B) * A) * B) * A), 1), 2);

optimized = 1 * (sum((((((((((sum(A, 1) * B) * A) * B) * A) * B) * A) * B) * A) * B) * A), 2));
normalization = sum(abs(original(:)));
assert(sum(abs(original(:) - optimized(:))) / normalization < 1e-10);
```

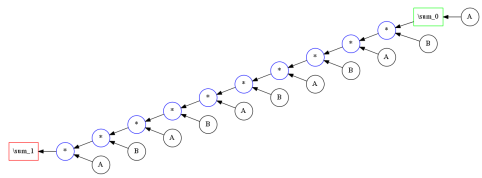

### k = 12

```
n = 100;
m = 200;
A = randn(n, m);
B = randn(m, n);
original = sum(sum(((((((((((A * B) * A) * B) * A) * B) * A) * B) * A) * B) * A) * B), 1), 2);

optimized = 1 * (sum(((((((((((sum(A, 1) * B) * A) * B) * A) * B) * A) * B) * A) * B) * A) * B), 2));
normalization = sum(abs(original(:)));
assert(sum(abs(original(:) - optimized(:))) / normalization < 1e-10);
```

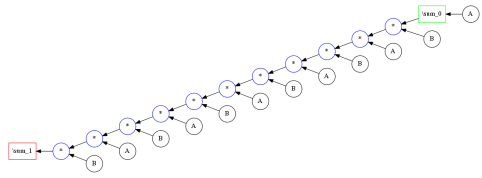

### k = 13

```
n = 100;
m = 200;
A = randn(n, m);
B = randn(m, n);
original = sum(sum((((((((((((A * B) * A) * B) * A) * B) * A) * B) * A) * B) * A) * B) * A), 1), 2);

optimized = 1 * (sum((((((((((((sum(A, 1) * B) * A) * B) * A) * B) * A) * B) * A) * B) * A) * B) * A), 2));
normalization = sum(abs(original(:)));
assert(sum(abs(original(:) - optimized(:))) / normalization < 1e-10);
```

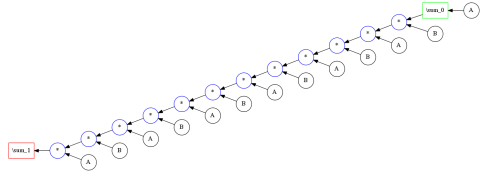

**k = 14**

```
n = 100;
m = 200;
A = randn(n, m);
B = randn(m, n);
original = sum(sum(((((((((((((A * B) * A) * B) * A) * B) * A) * B) * A) * B) * A) * B) * A) * B), 1), 2);

optimized = 1 * (sum(((((((((((((sum(A, 1) * B) * A) * B) * A) * B) * A) * B) * A) * B) * A) * B) * A) * B),
    ↪ 2));
normalization = sum(abs(original(:)));
assert(sum(abs(original(:) - optimized(:))) / normalization < 1e-10);
```

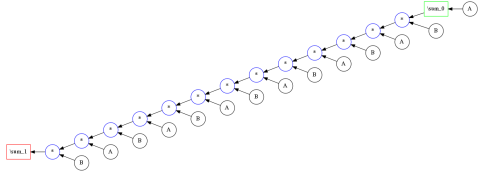

**k = 15**

```
n = 100;
m = 200;
A = randn(n, m);
B = randn(m, n);
original = sum(sum((((((((((((((A * B) * A) * B) * A) * B) * A) * B) * A) * B) * A) * B) * A) * B) * A), 1),
    ↪ 2);

optimized = 1 * (sum((((((((((((((sum(A, 1) * B) * A) * B) * A) * B) * A) * B) * A) * B) * A) * B) * A) * B)
    ↪ * A), 2));
normalization = sum(abs(original(:)));
assert(sum(abs(original(:) - optimized(:))) / normalization < 1e-10);
```

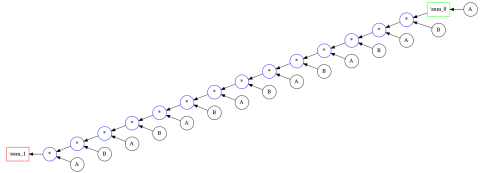

## 9.3 $\left(\sum (\mathbf{A}.*\mathbf{A})\mathbf{A}^{\mathbf{T}}\right)_k$

**k = 1**

```
n = 100;
m = 200;
A = randn(n, m);
original = sum(sum(A, 1), 2);

optimized = 1 * (sum(sum(A, 1)', 1));
normalization = sum(abs(original(:)));
assert(sum(abs(original(:) - optimized(:))) / normalization < 1e-10);
```

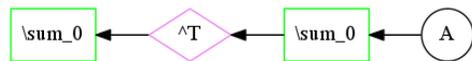

**k = 2**

```
n = 100;
m = 200;
A = randn(n, m);
original = sum(sum((A * A'), 1), 2);

optimized = 1 * ((sum(A, 1) * sum(A, 1)'));
normalization = sum(abs(original(:)));
assert(sum(abs(original(:) - optimized(:))) / normalization < 1e-10);
```

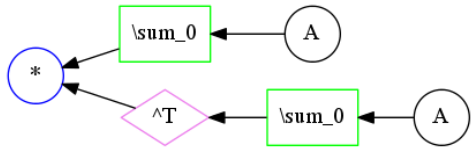

**k = 3**

```
n = 100;
m = 200;
A = randn(n, m);
original = sum(sum(((A * A') * (A .* A)), 1), 2);

optimized = 1 * (sum((sum((A' .* A'), 1) * (repmat(sum(A, 1)', 1, n) .* A')')', 1));
normalization = sum(abs(original(:)));
assert(sum(abs(original(:) - optimized(:))) / normalization < 1e-10);
```

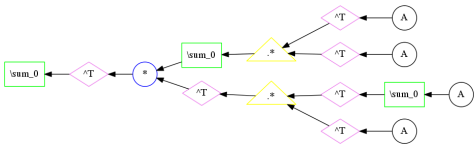

**k = 4**

```
n = 100;
m = 200;
A = randn(n, m);
original = sum(sum((((A * A') * (A .* A)) * A'), 1), 2);

optimized = 1 * (sum(sum((repmat(sum(A, 1)', 1, n) .* ((A' .* A')' .* repmat(sum((repmat(sum(A, 1)', 1, n) .*
    A'), 1)', 1, m))'), 1)', 1));
normalization = sum(abs(original(:)));
assert(sum(abs(original(:) - optimized(:))) / normalization < 1e-10);
```

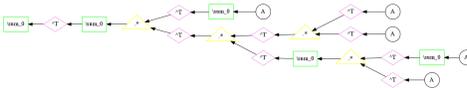

**k = 5**

```
n = 100;
m = 200;
A = randn(n, m);
original = sum(sum((((((A * A') * (A .* A)) * A') * (A .* A)), 1), 2);

optimized = 1 * (sum((sum((A' .* A'), 1) * (A' .* repmat((sum((repmat(sum(A, 1)', 1, n) .* A'), 1) * (A' .* A
    ')')', 1, n))')', 1));
normalization = sum(abs(original(:)));
assert(sum(abs(original(:) - optimized(:))) / normalization < 1e-10);
```

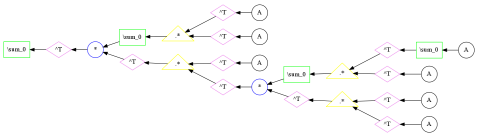

**k = 6**

```
n = 100;
m = 200;
A = randn(n, m);
original = sum(sum((((((A * A') * (A .* A)) * A') * (A .* A)) * A'), 1), 2);

optimized = 1 * (sum(sum((repmat(sum(A, 1)', 1, n) .* ((A' .* A')' .* repmat(sum((A' .* repmat((sum((repmat(
    sum(A, 1)', 1, n) .* A'), 1) * (A' .* A')')', 1, n)), 1)', 1, m))'), 1)', 1));
normalization = sum(abs(original(:)));
assert(sum(abs(original(:) - optimized(:))) / normalization < 1e-10);
```

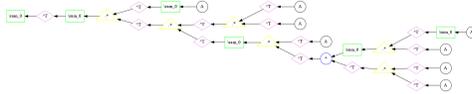

**k = 7**

```
n = 100;
m = 200;
A = randn(n, m);
original = sum(sum((((((((A * A') * (A .* A)) * A') * (A .* A)) * A') * (A .* A)), 1), 2);

optimized = 1 * (sum((sum((A' .* A'), 1) * (A' .* repmat((sum((A' .* repmat((sum((repmat(sum(A, 1)', 1, n) .*
    A'), 1) * (A' .* A')')', 1, n)), 1) * (A' .* A')')', 1, n))')', 1));
normalization = sum(abs(original(:)));
assert(sum(abs(original(:) - optimized(:))) / normalization < 1e-10);
```

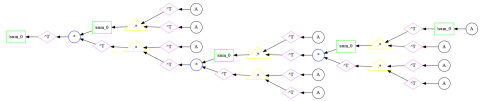

**k = 8**

```
n = 100;
m = 200;
A = randn(n, m);
original = sum(sum(((((((((A * A') * (A .* A)) * A') * (A .* A)) * A') * (A .* A)) * A'), 1), 2);

optimized = 1 * (sum(sum((repmat(sum(A, 1)', 1, n) .* ((A' .* A')' .* repmat(sum((A' .* repmat((sum((A' .*
    repmat((sum((repmat(sum(A, 1)', 1, n) .* A'), 1) * (A' .* A')')', 1, n)), 1) * (A' .* A')')', 1, n)),
    1)', 1, m))'), 1)', 1));
normalization = sum(abs(original(:)));
assert(sum(abs(original(:) - optimized(:))) / normalization < 1e-10);
```

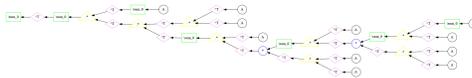

**k = 9**

```
n = 100;
m = 200;
A = randn(n, m);
original = sum(sum((((((((((A * A') * (A .* A)) * A') * (A .* A)) * A') * (A .* A)) * A') * (A .* A)), 1), 2);

optimized = 1 * (sum((sum((A' .* repmat((sum((A' .* repmat((sum((A' .* repmat((sum((repmat(sum(A, 1)', 1, n)
    .* A'), 1) * (A' .* A')')', 1, n)), 1) * (A' .* A')')', 1, n)), 1) * (A' .* A')')', 1, n)), 1) * (A'
    .* A')')', 1));
normalization = sum(abs(original(:)));
assert(sum(abs(original(:) - optimized(:))) / normalization < 1e-10);
```

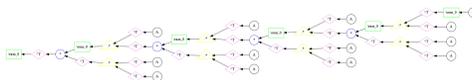

**k = 10**

```
n = 100;
m = 200;
A = randn(n, m);
original = sum(sum((((((((((((A * A') * (A .* A)) * A') * (A .* A)) * A') * (A .* A)) * A') * (A .* A)) * A'),
    1), 2);

optimized = 1 * (sum(sum((repmat(sum(A, 1)', 1, n) .* ((A' .* A')' .* repmat(sum((A' .* repmat((sum((A' .*
    repmat((sum((A' .* repmat((sum((repmat(sum(A, 1)', 1, n) .* A'), 1) * (A' .* A')')', 1, n)), 1) * (A'
    .* A')')', 1, n)), 1) * (A' .* A')')', 1, n)), 1)', 1, m))'), 1)', 1));
normalization = sum(abs(original(:)));
assert(sum(abs(original(:) - optimized(:))) / normalization < 1e-10);
```

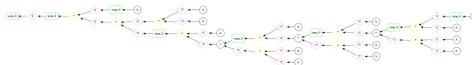

**k = 11**

```
n = 100;
m = 200;
A = randn(n, m);
original = sum(sum((((((((((A * A') * (A .* A)) * A') * (A .* A)) * A') * (A .* A)) * A') * (A .* A)) * A') *
    (A .* A)), 1), 2);

optimized = 1 * (sum((sum((A' .* repmat((sum((A' .* repmat((sum((A' .* repmat((sum((A' .* repmat((sum((repmat(
    sum(A, 1)', 1, n) .* A'), 1) * (A' .* A')')', 1, n)), 1) * (A' .* A')')', 1, n)), 1) * (A' .* A')')',
    1, n)), 1) * (A' .* A')')', 1, n)), 1) * (A' .* A')')', 1));
normalization = sum(abs(original(:)));
assert(sum(abs(original(:) - optimized(:))) / normalization < 1e-10);
```

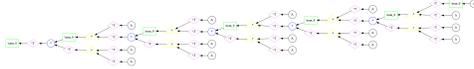

**k = 12**

```
n = 100;
m = 200;
A = randn(n, m);
original = sum(sum((((((((((((A * A') * (A .* A)) * A') * (A .* A)) * A') * (A .* A)) * A') * (A .* A)) * A')
    * (A .* A)) * A'), 1), 2);

optimized = 1 * (sum(sum((A' .* repmat((sum((A' .* repmat((sum((A' .* repmat((sum((A' .* repmat((sum((A' .*
    repmat((sum((repmat(sum(A, 1)', 1, n) .* A'), 1) * (A' .* A')')', 1, n)), 1) * (A' .* A')')', 1, n)),
    1) * (A' .* A')')', 1, n)), 1) * (A' .* A')')', 1, n)), 1) * (A' .* A')')', 1, n)), 1)', 1));
normalization = sum(abs(original(:)));
assert(sum(abs(original(:) - optimized(:))) / normalization < 1e-10);
```

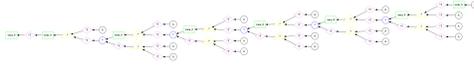

**k = 13**

```
n = 100;
m = 200;
A = randn(n, m);
original = sum(sum((((((((((((((A * A') * (A .* A)) * A') * (A .* A)) * A') * (A .* A)) * A') * (A .* A)) * A')
    * (A .* A)) * A') * (A .* A)), 1), 2);

optimized = 1 * (sum((sum((A' .* repmat((sum((A' .* repmat((sum((A' .* repmat((sum((A' .* repmat((sum((A' .*
    repmat((sum((repmat(sum(A, 1)', 1, n) .* A'), 1) * (A' .* A')')', 1, n)), 1) * (A' .* A')')', 1, n)),
    1) * (A' .* A')')', 1, n)), 1) * (A' .* A')')', 1, n)), 1) * (A' .* A')')', 1, n)), 1) * (A' .* A')
    ')', 1));
normalization = sum(abs(original(:)));
assert(sum(abs(original(:) - optimized(:))) / normalization < 1e-10);
```

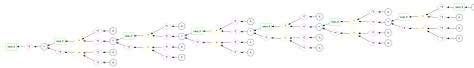

**k = 14**

```
n = 100;
m = 200;
A = randn(n, m);
original = sum(sum(((((((((((((((A * A') * (A .* A)) * A') * (A .* A)) * A') * (A .* A)) * A') * (A .* A)) * A
    ') * (A .* A)) * A') * (A .* A)) * A'), 1), 2);

optimized = 1 * (sum(sum((A' .* repmat((sum((A' .* repmat((sum((A' .* repmat((sum((A' .* repmat((sum((A' .*
    repmat((sum((A' .* repmat((sum((repmat(sum(A, 1)', 1, n) .* A'), 1) * (A' .* A')')', 1, n)), 1) * (A'
    .* A')')', 1, n)), 1) * (A' .* A')')', 1, n)), 1) * (A' .* A')')', 1, n)), 1) * (A' .* A')')', 1, n)
    ), 1) * (A' .* A')')', 1, n)), 1)', 1));
normalization = sum(abs(original(:)));
assert(sum(abs(original(:) - optimized(:))) / normalization < 1e-10);
```

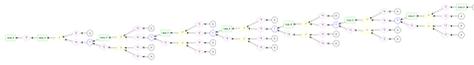

**k = 15**

```
n = 100;
m = 200;
A = randn(n, m);
original = sum(sum((((((((((((((A * A') * (A .* A)) * A') * (A .* A)) * A') * (A .* A)) * A
    ↪ ') * (A .* A)) * A') * (A .* A)) * A') * (A .* A)), 1), 2);

optimized = 1 * (sum((sum((A' .* repmat((sum((A' .* repmat((sum((A' .* repmat((sum((A' .* repmat((sum((A' .*
    ↪ repmat((sum((A' .* repmat((sum((repmat(sum(A, 1)', 1, n) .* A'), 1) * (A' .* A')')', 1, n)), 1) * (A'
    ↪ .* A')')', 1, n)), 1) * (A' .* A')')', 1, n)), 1) * (A' .* A')')', 1, n)), 1) * (A' .* A')')', 1, n)
    ↪ ), 1) * (A' .* A')')', 1, n)), 1) * (A' .* A')')', 1));
normalization = sum(abs(original(:)));
assert(sum(abs(original(:) - optimized(:))) / normalization < 1e-10);
```

### 9.4  Sym$_k$

**k = 1**

```
n = 1;
m = 18;
A = randn(1, m);
sub = nchoosek(1:m, 1);
original = 0;
for i = 1:size(sub, 1)
  original = original + prod(A(sub(i, :)));
end

optimized = (120 * (sum(A', 1))) / 120;
normalization = sum(abs(original(:)));
assert(sum(abs(original(:) - optimized(:))) / normalization < 1e-10);
```

**k = 2**

```
n = 1;
m = 18;
A = randn(1, m);
sub = nchoosek(1:m, 2);
original = 0;
for i = 1:size(sub, 1)
  original = original + prod(A(sub(i, :)));
end

optimized = (60 * (sum((repmat(sum(A', 1), 1, m) .* A), 2)) + -60 * (sum((A .* A)', 1))) / 120;
normalization = sum(abs(original(:)));
assert(sum(abs(original(:) - optimized(:))) / normalization < 1e-10);
```

**k = 3**

```
n = 1;
m = 18;
A = randn(1, m);
sub = nchoosek(1:m, 3);
original = 0;
for i = 1:size(sub, 1)
  original = original + prod(A(sub(i, :)));
end

optimized = (-60 * (sum((repmat(sum((A' .* A'), 1), 1, m) .* A), 2)) + 40 * (sum(((A' .* A')' .* A), 2)) + 20
    ↪ * (sum((A' .* repmat(sum(repmat(sum(A', 1), 1, m) .* A), 2), m, 1)), 1))) / 120;
normalization = sum(abs(original(:)));
assert(sum(abs(original(:) - optimized(:))) / normalization < 1e-10);
```

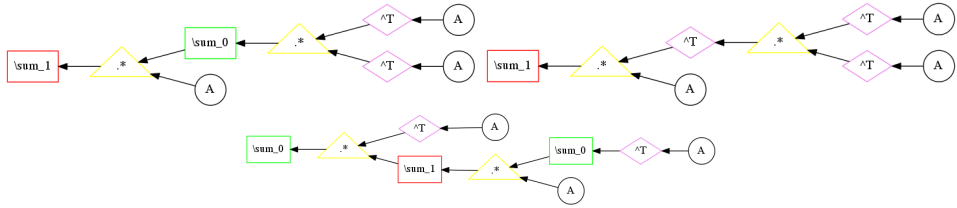

**k = 4**

```
n = 1;
m = 18;
A = randn(1, m);
sub = nchoosek(1:m, 4);
original = 0;
for i = 1:size(sub, 1)
  original = original + prod(A(sub(i, :)));
end

optimized = (5 * (sum((repmat(sum((A' .* repmat(sum((repmat(sum(A', 1), 1, m) .* A), 2), m, 1)), 1), 1, m) .*
    A), 2)) + −30 * (sum((((A' .* A')' .* A')' .* A'), 1)) + 15 * (sum(((repmat(sum((A' .* A'), 1), 1, m)
    .* A)' .* A'), 1)) + 40 * (sum((A' .* repmat(sum(((A' .* A')' .* A), 2), m, 1)), 1)) + −30 * (sum((A'
    .* repmat(sum((repmat(sum((A' .* A'), 1), 1, m) .* A), 2), m, 1)), 1))) / 120;
normalization = sum(abs(original(:)));
assert(sum(abs(original(:) − optimized(:))) / normalization < 1e−10);
```

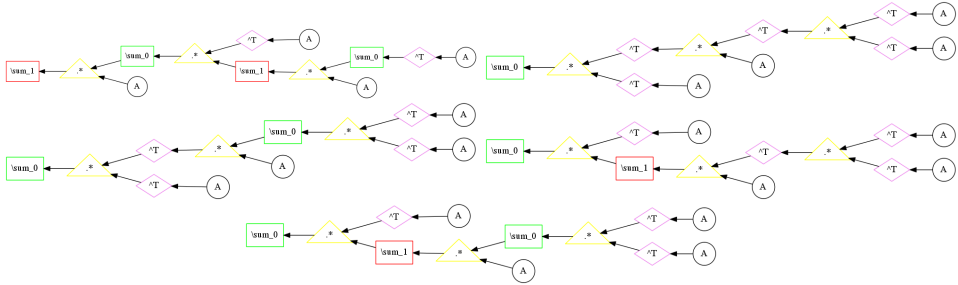

**k = 5**

```
n = 1;
m = 18;
A = randn(1, m);
sub = nchoosek(1:m, 5);
original = 0;
for i = 1:size(sub, 1)
  original = original + prod(A(sub(i, :)));
end

optimized = (−10 * (sum((repmat(sum((A' .* repmat(sum((repmat(sum((A' .* A'), 1), 1, m) .* A), 2), m, 1)), 1),
    1, m) .* A), 2)) + −20 * (sum((((repmat(sum((A' .* A'), 1), 1, m) .* A)' .* A')' .* A'), 1)) + 15 * (
    sum((repmat(sum((((repmat(sum((A' .* A'), 1), 1, m) .* A)' .* A')', 1), 1, m) .* A), 2)) + 24 * (sum
    (((((A' .* A')' .* A')' .* A')' .* A), 2)) + 20 * (sum((repmat(sum((A' .* repmat(sum(((A' .* A')' .* A
    ), 2), m, 1)), 1), 1, m) .* A), 2)) + 1 * (sum((A' .* repmat(sum((repmat(sum((A' .* repmat(sum((
    repmat(sum(A', 1), 1, m) .* A), 2), m, 1)), 1), 1, m) .* A), 2), m, 1)), 1)) + −30 * (sum((repmat(sum
    (((((A' .* A')' .* A)' .* A'), 1), 1, m) .* A), 2))) / 120;
normalization = sum(abs(original(:)));
assert(sum(abs(original(:) − optimized(:))) / normalization < 1e−10);
```

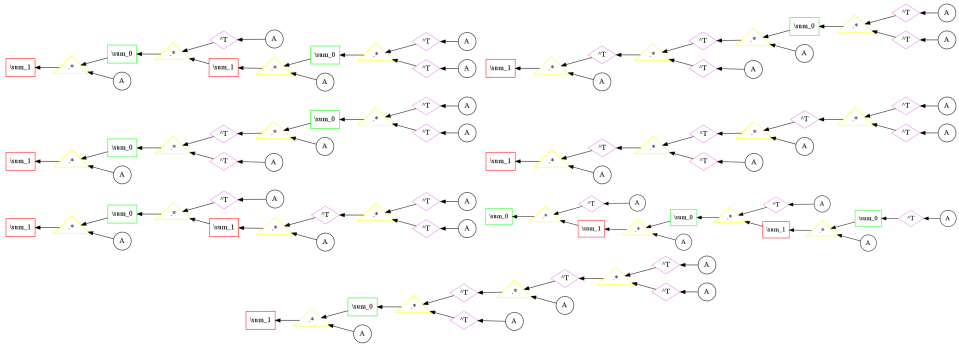

**k = 6**

```
n = 1;
m = 18;
A = randn(1, m);
sub = nchoosek(1:m, 6);
original = 0;
for i = 1:size(sub, 1)
  original = original + prod(A(sub(i, :)));
end

optimized = (24 * (sum((A' .* repmat(sum(((((A' .* A)' .* A)' .* A)' .* A), 2), m, 1)), 1)) + -5 / 2 * (sum
    ↪ (((repmat(sum(((repmat(sum((A' .* A), 1), 1, m) .* A)' .* A'), 1), 1, m) .* A)' .* A'), 1)) + 20 / 3
    ↪ * (sum((((A' .* repmat(sum(((A' .* A)' .* A), 2), m, 1))' .* A')' .* A'), 1)) + -20 * (sum(((((((A'
    ↪ .* A)' .* A)' .* A)' .* A)' .* A'), 1)) + 15 * (sum(((repmat(sum((((A' .* A)' .* A)' .* A'), 1),
    ↪ 1, m) .* A)' .* A'), 1)) + -20 * (sum((A' .* repmat(sum((((repmat(sum((A' .* A), 1), 1, m) .* A)' .*
    ↪ A)' .* A), 2), m, 1)), 1)) + -15 * (sum((A' .* repmat(sum((repmat(sum((((A' .* A)' .* A)' .* A'),
    ↪ 1), 1, m) .* A)' .* A'), 2), m, 1)), 1)) + 15 / 2 * (sum((A' .* repmat(sum((repmat(sum(((A' .* A
    ↪ ')' .* A'), 1), 1, m) .* A)' .* A'), 1), 1, m) .* A), 2), m, 1)), 1)) + 1 / 6 * (sum((repmat(sum((A' .*
    ↪ repmat(sum((repmat(sum((A' .* repmat(sum((repmat(sum(A', 1), 1, m) .* A), 2), m, 1)), 1), 1, m) .* A)
    ↪ , 2), m, 1)), 1), 1, m) .* A), 2)) + 20 / 3 * (sum((A' .* repmat(sum((A' .* repmat(sum(((
    ↪ A' .* A)' .* A), 2), m, 1)), 1), 1, m) .* A), 2), m, 1)), 1)) + -5 / 2 * (sum((A' .* repmat(sum((
    ↪ repmat(sum((A' .* repmat(sum((repmat(sum((A' .* A'), 1), 1, m) .* A), 2), m, 1)), 1), 1, m) .* A), 2)
    ↪ , m, 1)), 1))) / 120;
normalization = sum(abs(original(:)));
assert(sum(abs(original(:) - optimized(:))) / normalization < 1e-10);
```

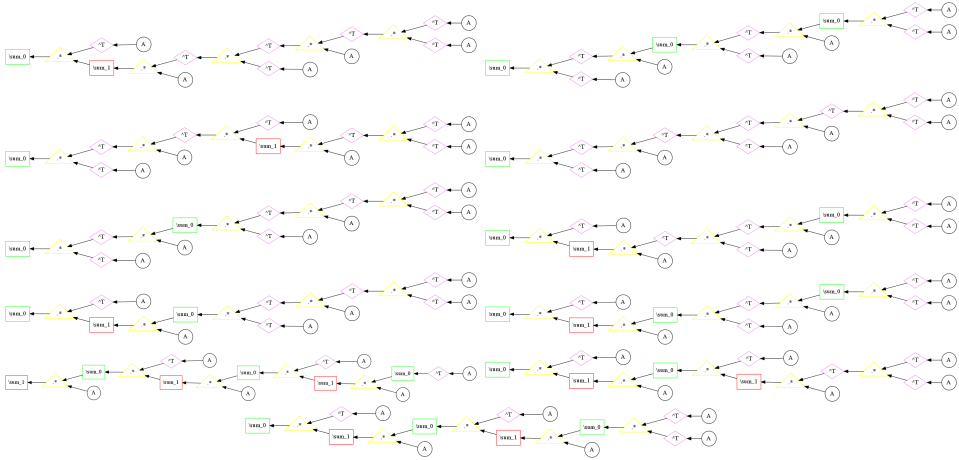

**k = 7**

```
n = 1;
m = 18;
A = randn(1, m);
sub = nchoosek(1:m, 7);
original = 0;
for i = 1:size(sub, 1)
  original = original + prod(A(sub(i, :)));
end

optimized = (20 / 3 * (sum((repmat(sum((((A' .* repmat(sum(((A' .* A')' .* A), 2), m, 1))' .* A)' .* A'), 1),
    ↪ 1, m) .* A), 2)) + -10 * (sum((repmat(sum((A' .* repmat(sum(((A' .* A'), 1), 1, m) .* A)
    ↪ ' .* A')' .* A), 2), m, 1)), 1), 1, m) .* A), 2)) + 5 / 3 * (sum((repmat(sum((A' .* repmat(sum((
    ↪ repmat(sum((A' .* repmat(sum(((A' .* A')' .* A), 2), m, 1)), 1), 1, m) .* A), 2), m, 1)), 1), 1, m)
    ↪ .* A), 2)) + -5 / 2 * (sum((repmat(sum(((repmat(sum((A' .* repmat(sum(((A' .* A')' .* A'),
    ↪ 1), 1, m) .* A)' .* A'), 1), 1, m) .* A), 2)) + 15 * (sum((repmat(sum((((repmat(sum(((((A' .* A')' .*
    ↪ A)' .* A'), 1), 1, m) .* A)' .* A'), 1), 1, m) .* A), 2)) + -12 * (sum(((((repmat(sum((A' .* A'), 1)
    ↪ , 1, m) .* A)' .* A')' .* A)' .* A'), 2)) + 5 / 2 * (sum((repmat(sum((A' .* repmat(sum((repmat
    ↪ (sum(((repmat(sum((A' .* A'), 1), 1, m) .* A)' .* A'), 1), 1, m) .* A), 2), m, 1)), 1), 1, m) .* A),
    ↪ 2)) + 1 / 42 * (sum((A' .* repmat(sum((repmat(sum((A' .* repmat(sum((repmat(sum((A' .* repmat(sum((
    ↪ repmat(sum(A', 1), 1, m) .* A), 2), m, 1)), 1), 1, m) .* A), 2), m, 1)), 1), 1, m) .* A), 2), m, 1)),
    ↪ 1)) + -5 * (sum((repmat(sum((A' .* repmat(sum((repmat(sum(((((A' .* A')' .* A')' .* A'), 1), 1, m) .*
    ↪ A), 2), m, 1)), 1), 1, m) .* A), 2)) + 5 * (sum(((((repmat(sum(((repmat(sum((A' .* A'), 1), 1, m) .* A
    ↪ )' .* A'), 1), 1, m) .* A)' .* A')' .* A'), 2)) + 120 / 7 * (sum((((((((A' .* A')' .* A)' .* A')'
    ↪ .* A')' .* A), 2)) + -10 * (sum(((((repmat(sum((((A' .* A')' .* A')' .* A'), 1), 1, m) .* A)' .* A')'
    ↪ .* A), 2)) + 12 * (sum((repmat(sum((A' .* repmat(sum(((((A' .* A')' .* A)' .* A)' .* A), 2), m, 1))
    ↪ , 1), 1, m) .* A), 2)) + -1 / 2 * (sum((repmat(sum((A' .* repmat(sum((repmat(sum((A' .* repmat(sum((
    ↪ repmat(sum((A' .* A'), 1), 1, m) .* A), 2), m, 1)), 1), 1, m) .* A), 2), m, 1)), 1), 1, m) .* A), 2))
    ↪ + -20 * (sum((repmat(sum((((((A' .* A')' .* A')' .* A')' .* A'), 1), 1, m) .* A), 2))) / 120;
normalization = sum(abs(original(:)));
assert(sum(abs(original(:) - optimized(:))) / normalization < 1e-10);
```

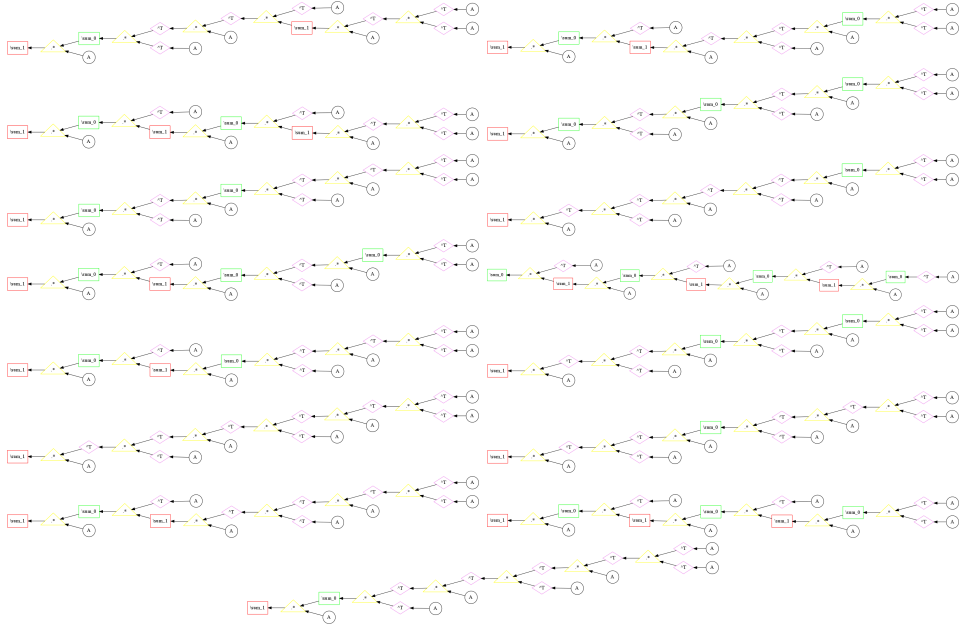

**k = 8**

```
n = 1;
m = 18;
A = randn(1, m);
sub = nchoosek(1:m, 8);
original = 0;
for i = 1:size(sub, 1)
    original = original + prod(A(sub(i, :)));
end

optimized = (5 / 8 * (sum((A' .* repmat(sum((repmat(sum((A' .* repmat(sum((repmat(sum(((repmat(sum((A' .* A'),
    1), 1, m) .* A)' .* A'), 1), 1, m) .* A), 2), m, 1)), 1), 1, m) .* A), 2), m, 1)), 1)) + -5 / 4 * (
    sum((A' .* repmat(sum((repmat(sum(((repmat(sum((A' .* A'), 1), 1, m) .* A)' .* A'), 1),
    1, m) .* A)' .* A'), 1), 1, m) .* A), 2), m, 1)), 1)) + 4 * (sum((A' .* repmat(sum((repmat(sum((A' .*
    repmat(sum(((((A' .* A')' .* A')' .* A')' .* A), 2), m, 1)), 1), 1, m) .* A), 2), m, 1)), 1)) + -5 /
    4 * (sum((A' .* repmat(sum((repmat(sum((A' .* repmat(sum((((A' .* A')' .* A')' .* A'), 1),
    1, m) .* A), 2), m, 1)), 1), 1, m) .* A), 2), m, 1)), 1)) + 5 / 16 * (sum(((repmat(sum(((repmat(sum
    (((repmat(sum((A' .* A'), 1), 1, m) .* A)' .* A'), 1), 1, m) .* A)' .* A'), 1), 1, m) .* A)' .* A'),
    1)) + 15 / 2 * (sum((A' .* repmat(sum(((repmat(sum((((A' .* A')' .* A')' .* A'), 1), 1, m)
    .* A)' .* A'), 1), 1, m) .* A), 2), m, 1)), 1)) + 8 * (sum((((A' .* repmat(sum(((((A' .* A')' .* A)'
    .* A')' .* A), 2), m, 1))' .* A')' .* A'), 1)) + 1 / 3 * (sum((A' .* repmat(sum((repmat(sum((A' .*
    repmat(sum((repmat(sum((A' .* repmat(sum(((A' .* A'), 2), m, 1)), 1), 1, m) .* A), 2), m, 1)),
    1), 1, m) .* A), 2), m, 1)), 1)) + 120 / 7 * (sum((A' .* repmat(sum((((((A' .* A')' .* A')' .* A')'
    .* A)' .* A')' .* A), 2), m, 1)), 1)) + 5 * (sum((A' .* repmat(sum(((((repmat(sum(((repmat(sum((A' .*
    A'), 1), 1, m) .* A)' .* A'), 1), 1, m) .* A)' .* A'), 2), m, 1)), 1)) + -10 * (sum((A' .*
    repmat(sum((repmat(sum((((((A' .* A')' .* A')' .* A')' .* A')' .* A'), 1), 1, m) .* A), 2), m, 1)), 1))
    + -15 / 4 * (sum(((repmat(sum(((repmat(sum((((A' .* A')' .* A')' .* A'), 1), 1, m) .* A)' .* A'), 1),
    1, m) .* A)' .* A'), 1)) + 10 / 3 * (sum((A' .* repmat(sum(((A' .* repmat(sum(((A' .* A
    ')' .* A), 2), m, 1))' .* A)' .* A'), 1), 1, m) .* A), 2), m, 1)), 1)) + 15 / 4 * (sum(((((repmat(sum
    ((((A' .* A')' .* A')' .* A'), 1), 1, m) .* A)' .* A')' .* A')' .* A'), 1)) + -1 / 12 * (sum((A' .*
    repmat(sum((repmat(sum((A' .* repmat(sum((repmat(sum((A' .* repmat(sum((A' .* A'), 1), 1,
    m) .* A), 2), m, 1)), 1), 1, m) .* A), 2), m, 1)), 1), 1, m) .* A), 2), m, 1)), 1)) + -15 * (sum
    (((((((((A' .* A')' .* A')' .* A')' .* A')' .* A')' .* A')' .* A'), 1)) + -10 / 3 * (sum((A' .* repmat(
    sum((repmat(sum((A' .* repmat(sum((((repmat(sum((A' .* A'), 1), 1, m) .* A), 2), m, 1))' .* A)' .* A'),
    1), 1, m) .* A), 2), m, 1)), 1)) + -10 * (sum((A' .* repmat(sum(((((repmat(sum((((A' .* A')' .* A)'
    .* A'), 1), 1, m) .* A)' .* A')' .* A), 2), m, 1)), 1)) + 10 * (sum(((repmat(sum((((((A' .* A')' .*
    A)' .* A')' .* A'), 1), 1, m) .* A)' .* A')' .* A'), 1)) + -12 * (sum((A' .* repmat(sum((((((repmat(
    sum((A' .* A'), 1), 1, m) .* A)' .* A')' .* A')' .* A)' .* A), 2), m, 1)), 1)) + 1 / 336 * (sum((
    repmat(sum((A' .* repmat(sum((repmat(sum((A' .* repmat(sum((repmat(sum((A' .* repmat(sum((repmat(sum
    (A', 1), 1, m) .* A), 2), m, 1)), 1), 1, m) .* A), 2), m, 1)), 1), 1, m)
    .* A), 2)) + -10 / 3 * (sum((((A' .* repmat(sum(((((repmat(sum((A' .* A'), 1), 1, m) .* A)' .* A')' .*
    A), 2), m, 1))' .* A)' .* A'), 1))) / 120;
normalization = sum(abs(original(:)));
assert(sum(abs(original(:) - optimized(:))) / normalization < 1e-10);
```

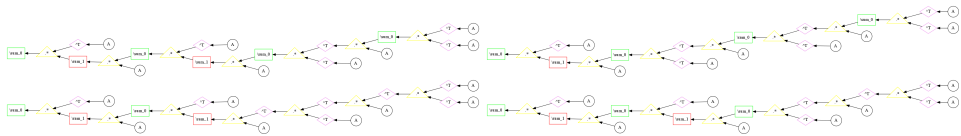

15

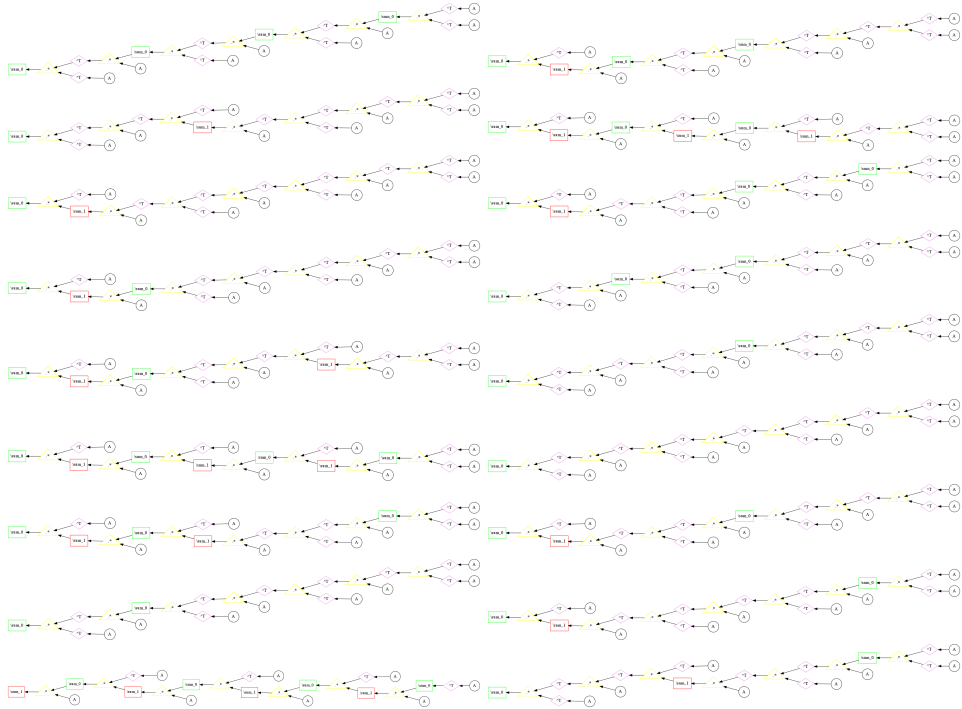

### 9.5 (RBM-1)$_k$

**k = 1**

```
n = 14;
m = 1;
A = randn(1, n);
nset = dec2bin(0:(2^(n) - 1));
original = 0;
for i = 1:size(nset, 1)
  v = logical(nset(i, :) - '0');
  original = original + (v * A') ^ 1;
end

optimized = 2^(n - 3) * (4 * (sum(A, 2)));
normalization = sum(abs(original(:)));
assert(sum(abs(original(:) - optimized(:))) / normalization < 1e-10);
```

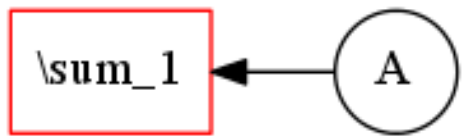

**k = 2**

```
n = 14;
m = 1;
A = randn(1, n);
nset = dec2bin(0:(2^(n) - 1));
original = 0;
for i = 1:size(nset, 1)
  v = logical(nset(i, :) - '0');
  original = original + (v * A') ^ 2;
end

optimized = 2^(n - 3) * (2 * (sum(sum((A' * A), 2)', 2)) + 2 * (sum((A .* A), 2)));
normalization = sum(abs(original(:)));
assert(sum(abs(original(:) - optimized(:))) / normalization < 1e-10);
```

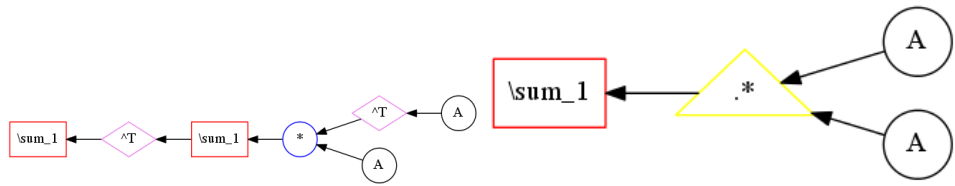

**k = 3**

```
n = 14;
m = 1;
A = randn(1, n);
nset = dec2bin(0:(2^(n) − 1));
original = 0;
for i = 1:size(nset, 1)
  v = logical(nset(i, :) − '0');
  original = original + (v * A') ^ 3;
end

optimized = 2^(n − 4) * (2 * (((sum(A, 2) * sum(A, 2)) * sum(A, 2))) + 6 * ((A * (sum(A, 2) * A)')));
normalization = sum(abs(original(:)));
assert(sum(abs(original(:) − optimized(:))) / normalization < 1e−10);
```

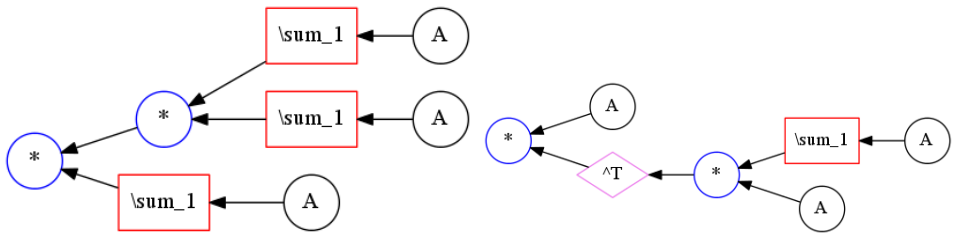

**k = 4**

```
n = 14;
m = 1;
A = randn(1, n);
nset = dec2bin(0:(2^(n) − 1));
original = 0;
for i = 1:size(nset, 1)
  v = logical(nset(i, :) − '0');
  original = original + (v * A') ^ 4;
end

optimized = 2^(n − 5) * (2 * ((((sum(A, 2) * sum(A, 2)) * sum(A, 2)) * sum(A, 2))) + −4 * (((A .* A) * (A .* A
    )')) + 6 * ((A * (sum((A .* A), 2) * A)')) + 12 * (((sum(A, 2) * sum(A, 2)) * sum((A .* A), 2))));
normalization = sum(abs(original(:)));
assert(sum(abs(original(:) − optimized(:))) / normalization < 1e−10);
```

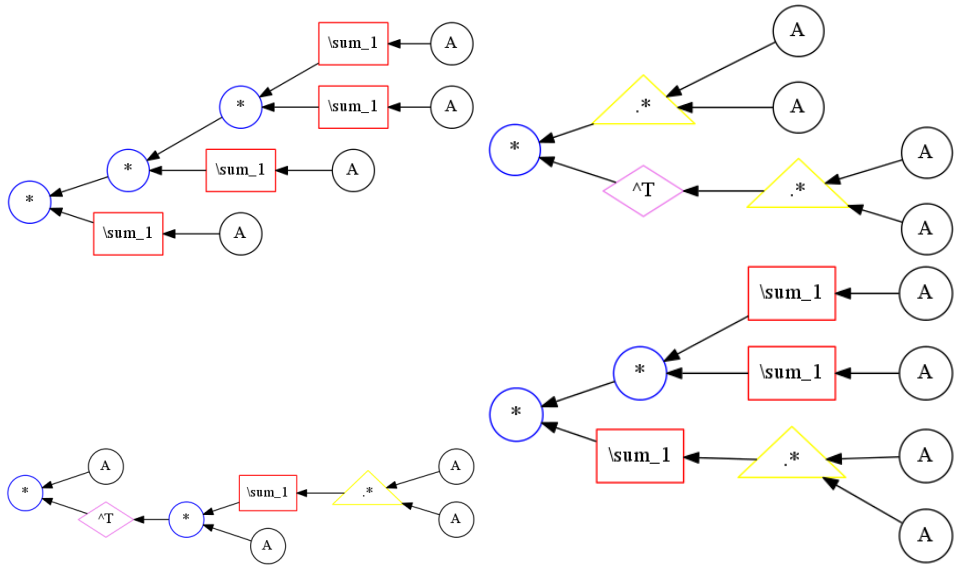

**k = 5**

```
n = 14;
m = 1;
A = randn(1, n);
nset = dec2bin(0:(2^(n) − 1));
original = 0;
for i = 1:size(nset, 1)
  v = logical(nset(i, :) − '0');
  original = original + (v * A') ^ 5;
end

optimized = 2^(n − 6) * (2 * (((((sum(A, 2) * sum(A, 2)) * sum(A, 2)) * sum(A, 2)) * sum(A, 2))) + 20 * ((((A
    * (sum(A, 2) * A)') * sum(A, 2)) * sum(A, 2))) + −20 * ((((A .* A) * (A .* A)') * sum(A, 2))) + 30 *
    (((A * (sum((A .* A), 2) * A)') * sum(A, 2))));
normalization = sum(abs(original(:)));
assert(sum(abs(original(:) − optimized(:))) / normalization < 1e−10);
```

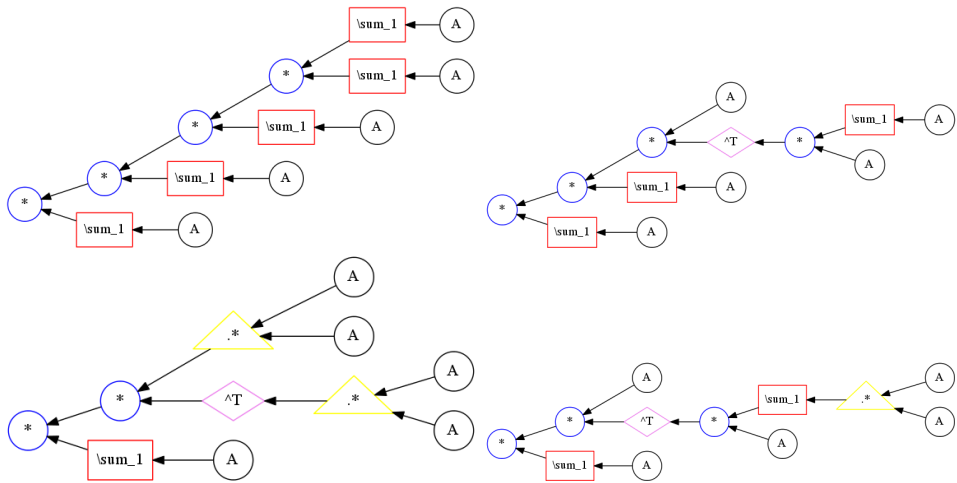

**k = 6**

```
n = 14;
m = 1;
A = randn(1, n);
nset = dec2bin(0:(2^(n) − 1));
original = 0;
for i = 1:size(nset, 1)
  v = logical(nset(i, :) − '0');
  original = original + (v * A') ^ 6;
end
```

```
optimized = 2^(n − 7) * (2 * ((((((sum(A, 2) * sum(A, 2)) * sum(A, 2)) * sum(A, 2)) * sum(A, 2)) * sum(A, 2)))
    + 30 * (((((A * (sum(A, 2) * A)') * sum(A, 2)) * sum(A, 2)) * sum(A, 2))) + 30 * (((A * (sum((A .* A
    ), 2) * A)') * sum((A .* A), 2))) + −60 * ((((A .* A) * (A .* A)') * sum((A .* A), 2))) + 90 * ((((A
    * (sum((A .* A), 2) * A)') * sum(A, 2)) * sum(A, 2))) + −60 * (((((A .* A) * (A .* A)') * sum(A, 2))
    * sum(A, 2))) + 32 * (((A' .* (A .* A)')' * (A' .* (A .* A)'))));
normalization = sum(abs(original(:)));
assert(sum(abs(original(:) − optimized(:))) / normalization < 1e−10);
```

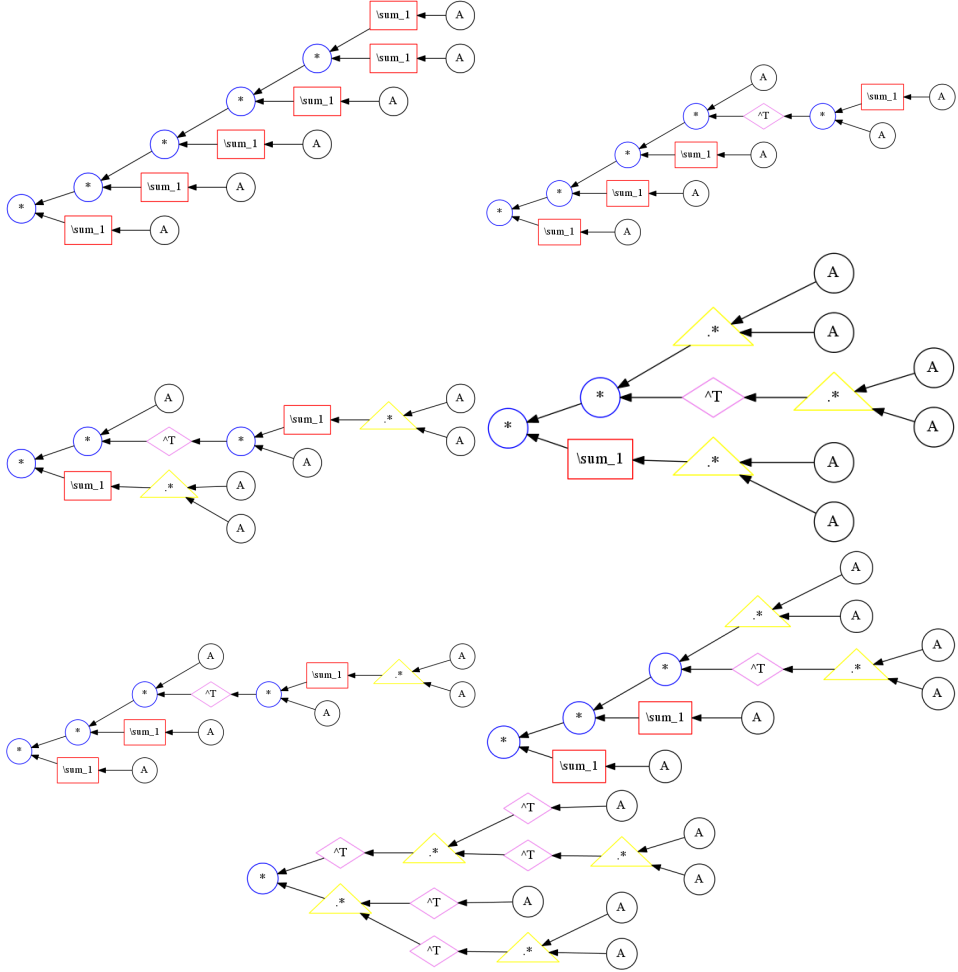

**k = 7**

```
n = 14;
m = 1;
A = randn(1, n);
nset = dec2bin(0:(2^(n) − 1));
original = 0;
for i = 1:size(nset, 1)
    v = logical(nset(i, :) − '0');
    original = original + (v * A') ^ 7;
end

optimized = 2^(n − 8) * (42 * ((((((A * (sum(A, 2) * A)') * sum(A, 2)) * sum(A, 2)) * sum(A, 2)) * sum(A, 2)))
    + −420 * (((((A .* A) * (A .* A)') * sum((A .* A), 2)) * sum(A, 2))) + −140 * (((((A .* A) * (A .*
    A)') * sum(A, 2)) * sum(A, 2)) * sum(A, 2))) + 2 * (((((((sum(A, 2) * sum(A, 2)) * sum(A, 2)) * sum(A
    , 2)) * sum(A, 2)) * sum(A, 2)) * sum(A, 2))) + 210 * ((((A * (sum((A .* A), 2) * A)') * sum((A .* A)
    , 2)) * sum(A, 2))) + 224 * ((((A' .* (A .* A)')' * (A' .* (A .* A)')) * sum(A, 2))) + 210 * (((((A *
    (sum((A .* A), 2) * A)') * sum(A, 2)) * sum(A, 2)) * sum(A, 2))));
normalization = sum(abs(original(:)));
assert(sum(abs(original(:) − optimized(:))) / normalization < 1e−10);
```

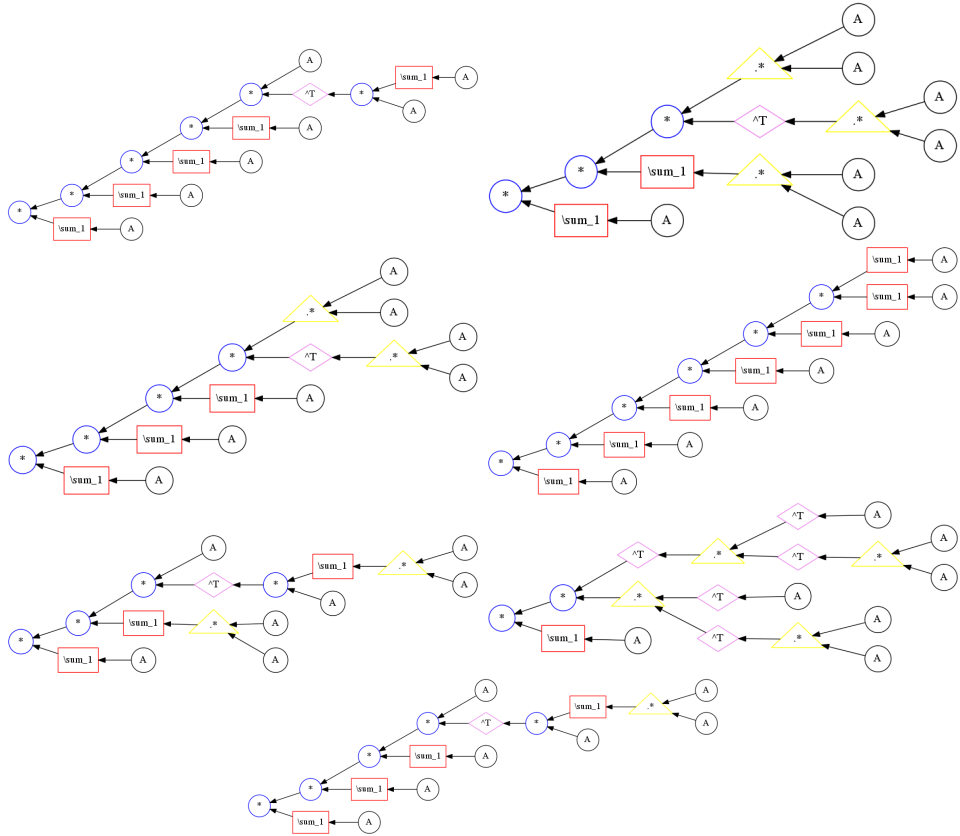

**k = 8**

```
n = 14;
m = 1;
A = randn(1, n);
nset = dec2bin(0:(2^(n) - 1));
original = 0;
for i = 1:size(nset, 1)
  v = logical(nset(i, :) - '0');
  original = original + (v * A') ^ 8;
end

optimized = 2^(n - 9) * (56 * (((((((A * (sum(A, 2) * A)') * sum(A, 2)) * sum(A, 2)) * sum(A, 2)) * sum(A, 2))
    ↪ * sum(A, 2))) + 2 * (((((((((sum(A, 2) * sum(A, 2)) * sum(A, 2)) * sum(A, 2)) * sum(A, 2)) * sum(A,
    ↪ 2)) * sum(A, 2)) * sum(A, 2))) + 210 * ((((A * (sum((A .* A), 2) * A)')' * sum((A .* A), 2)) * sum((A
    ↪ .* A), 2))) + 896 * (((((A' .* (A .* A)')' * (A' .* (A .* A)')) * sum(A, 2)) * sum(A, 2))) + 420 *
    ↪ (((((((A * (sum((A .* A), 2) * A)') * sum(A, 2)) * sum(A, 2)) * sum(A, 2)) * sum(A, 2))) + -544 * (((A
    ↪ .* A) * ((A' .* (A .* A)')' .* (A' .* (A .* A)')')'))) + 896 * ((((A' .* (A .* A)')' * (A' .* (A .* A
    ↪ )')) * sum((A .* A), 2))) + -1680 * ((((((A .* A) * (A .* A)') * sum((A .* A), 2)) * sum(A, 2)) * sum
    ↪ (A, 2))) + 280 * ((((A .* A) * (A .* A)') * ((A .* A) * (A .* A)'))) + 840 * (((((A * (sum((A .* A),
    ↪ 2) * A)') * sum((A .* A), 2)) * sum((A .* A), 2)) * sum(A, 2))) + -280 * (((((((A .* A) * (A .* A)') * sum(A
    ↪ , 2)) * sum(A, 2)) * sum(A, 2)) * sum(A, 2))) + -840 * (((((A .* A) * (A .* A)') * sum((A .* A), 2))
    ↪ * sum((A .* A), 2))));
normalization = sum(abs(original(:)));
assert(sum(abs(original(:) - optimized(:))) / normalization < 1e-10);
```

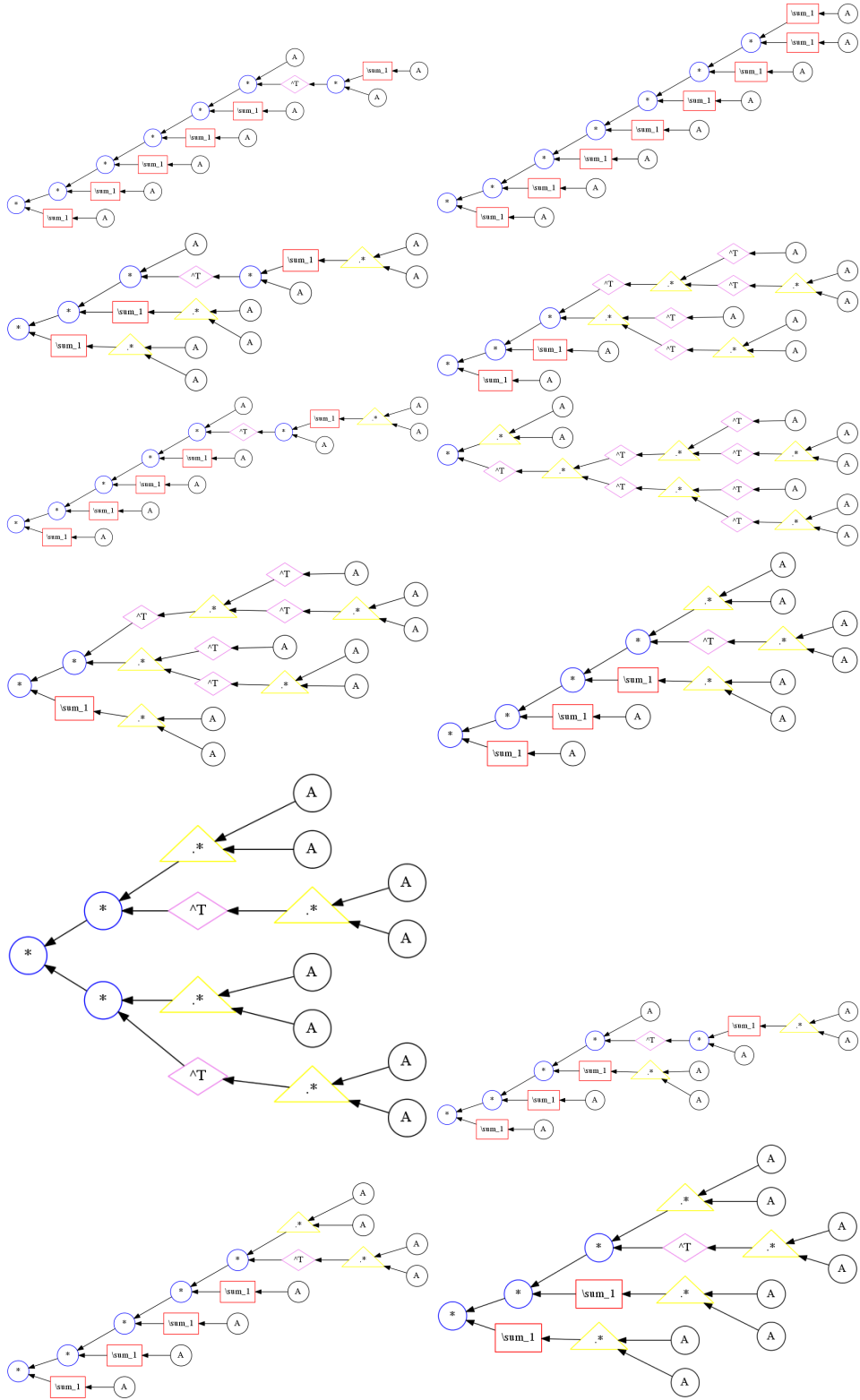

## 9.6 (RBM-2)$_k$

**k = 1**

```
n = 7;
m = 8;
```

```
A = randn(n, m);
nset = dec2bin(0:(2^(n) - 1));
mset = dec2bin(0:(2^(m) - 1));
original = 0;
for i = 1:size(nset, 1)
  for j = 1:size(mset, 1)
    v = logical(nset(i, :) - '0');
    h = logical(mset(j, :) - '0');
    original = original + (v * A * h') ^ 1;
  end
end

optimized = 2^(n + m - 5) * (8 * (sum(sum(A', 1)', 1)));
normalization = sum(abs(original(:)));
assert(sum(abs(original(:) - optimized(:))) / normalization < 1e-10);
```

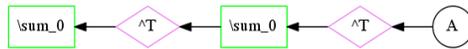

**k = 2**

```
n = 7;
m = 8;
A = randn(n, m);
nset = dec2bin(0:(2^(n) - 1));
mset = dec2bin(0:(2^(m) - 1));
original = 0;
for i = 1:size(nset, 1)
  for j = 1:size(mset, 1)
    v = logical(nset(i, :) - '0');
    h = logical(mset(j, :) - '0');
    original = original + (v * A * h') ^ 2;
  end
end

optimized = 2^(n + m - 5) * (2 * (sum((sum(A, 1) .* sum(A, 1)), 2)) + 2 * (sum(sum((A .* A), 1), 2)) + 2 * (
    sum(sum((A .* repmat(sum(A, 2), 1, m)), 1)', 1)) + 2 * (sum((sum(A, 1) .* repmat(sum(sum(A, 1)', 1),
    1, m)), 2)));
normalization = sum(abs(original(:)));
assert(sum(abs(original(:) - optimized(:))) / normalization < 1e-10);
```

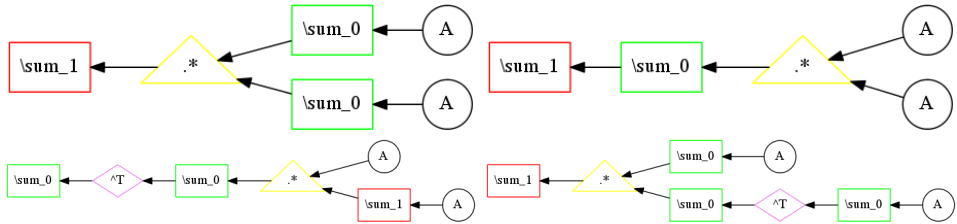

**k = 3**

```
n = 7;
m = 8;
A = randn(n, m);
nset = dec2bin(0:(2^(n) - 1));
mset = dec2bin(0:(2^(m) - 1));
original = 0;
for i = 1:size(nset, 1)
  for j = 1:size(mset, 1)
    v = logical(nset(i, :) - '0');
    h = logical(mset(j, :) - '0');
    original = original + (v * A * h') ^ 3;
  end
end

optimized = 2^(n + m - 7) * (12 * (sum(sum((A .* repmat(sum((A .* repmat(sum(A', 2)', n, 1)), 2), 1, m)), 1),
    2)) + 2 * (sum(sum((A .* repmat(sum((sum(A', 1) .* repmat(sum(sum(A', 1)', 1), 1, n)), 2), n, m)), 1)
    , 2)) + 6 * (sum(sum((A .* repmat(sum((sum(A', 1) .* sum(A', 1)), 2), n, m)), 1), 2)) + 6 * (sum(sum
    (((repmat(sum(sum(A', 1)', 1), n, m) .* A) .* A'), 1), 2)) + 6 * (sum(sum((A .* repmat(sum((repmat(
    sum(sum(A', 1)', 1), n, m) .* A), 1), n, 1)), 1), 2)));
normalization = sum(abs(original(:)));
assert(sum(abs(original(:) - optimized(:))) / normalization < 1e-10);
```

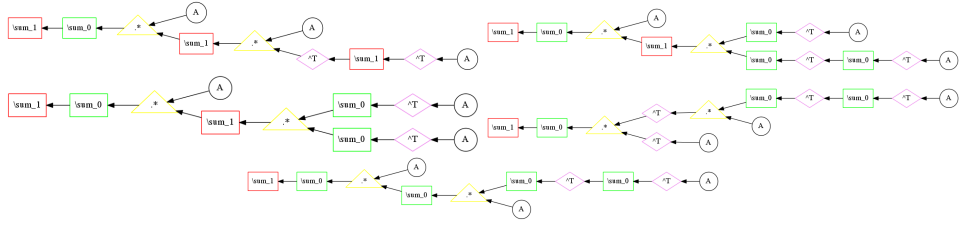

**k = 4**

```
n = 7;
m = 8;
A = randn(n, m);
nset = dec2bin(0:(2^(n) − 1));
mset = dec2bin(0:(2^(m) − 1));
original = 0;
for i = 1:size(nset, 1)
  for j = 1:size(mset, 1)
    v = logical(nset(i, :) − '0');
    h = logical(mset(j, :) − '0');
    original = original + (v * A * h') ^ 4;
  end
end
optimized = 2^(n + m − 9) * (24 * ((sum(A, 1) * sum((repmat(sum((repmat(sum(A, 1), n, 1) .* A)', 1), m, 1) .*
    A'), 2))) + 6 * (sum(sum((A .* repmat(sum((sum(A', 1) .* sum(A', 1)), 2), n, m)), 2
    , 1, m)), 1), 2) + −24 * (sum(sum((A .* repmat(sum((A .* repmat(sum((A .* A), 1), n, 1)), 1), n, 1))
    , 1), 2) + 8 * (sum(sum(((A'' .* (A .* A)) .* A), 1), 2)) + 12 * (sum(sum((A .* repmat(sum((A .*
    repmat(sum((A .* A), 1), 2), n, m)), 2), 1, m)), 1), 2) + −24 * (sum(sum((A .* repmat(sum((A .*
    repmat(sum((A .* A), 2), 1, m)), 2), 1, m)), 1), 2)) + 12 * (sum(sum(((A .* A') .* (A * A')), 1), 2))
    + 12 * (sum(sum((A .* repmat(sum((A .* repmat(sum((sum(A, 1) .* repmat(sum(sum(A, 1)', 1), 1, m)), 2)
    , n, m)), 2), 1, m)), 1), 2)) + −4 * (sum(sum((A .* repmat(sum((repmat(sum(A, 1), n, 1) .* (repmat(
    sum(A, 1), n, 1) .* A)), 1), n, 1)), 1), 2) + 24 * (sum(sum((A .* repmat(sum((A .*
    repmat(sum(A, 2), 1, m)), 1), n, 1)), 2), 1, m)), 1), 2)) + 2 * (sum(sum((A .* repmat(sum(sum((A .*
    repmat(sum((sum(A, 1) .* repmat(sum(sum(A, 1)', 1), 1, m)), 2), n, m)), 1), 2)) + 12
    * (sum((repmat(sum(sum(A, 1), 2), 1, m) .* (repmat(sum(sum(A, 1)', 1), 1, m) .* sum((A .* A), 1))),
    2)) + 12 * (sum(sum((A .* repmat(sum((A .* repmat(sum(sum((A .* A), 1), 2), n, m)), 1), n, 1)), 1),
    2)) + 6 * (sum(sum((A .* repmat(sum((A .* repmat(sum((sum(A, 1) .* sum(A, 1)), 2), n, m)), 1), n, 1))
    , 1), 2) + 6 * (sum((sum((A .* A), 2) .* repmat(sum(sum((A .* A), 1), 2), n, 1)), 1)) + 12 * (sum(
    sum((A .* repmat(sum((A .* repmat(sum((sum(A, 1) .* sum(A, 1)), 2), n, m)), 2), 1, m)), 1), 2)) + 48
    * (sum(sum((A .* repmat(sum(sum((A .* repmat(sum(A, 2), 1, m)), 1), n, 1)), 1), 2),
    n, m)), 1), 2) + −12 * (sum((sum((A .* A), 2) .* sum((A .* A), 2)), 1)) + −12 * (sum((sum((A .* A),
    1) .* sum((A .* A), 1)), 2)) + 12 * (sum(sum((A .* repmat(sum(sum((A .* repmat(sum((sum(A, 1) .* sum(
    A, 1)), 2), n, m)), 1), 2), n, m)), 1), 2), n, m)), 1), 2)) + −4 * (sum(sum((A .* repmat(sum((A .* repmat(sum((A .*
    repmat(sum(A, 2), 1, m)), 2), 1, m)), 2), 1, m)), 2)));
normalization = sum(abs(original(:)));
assert(sum(abs(original(:) − optimized(:))) / normalization < 1e−10);
```

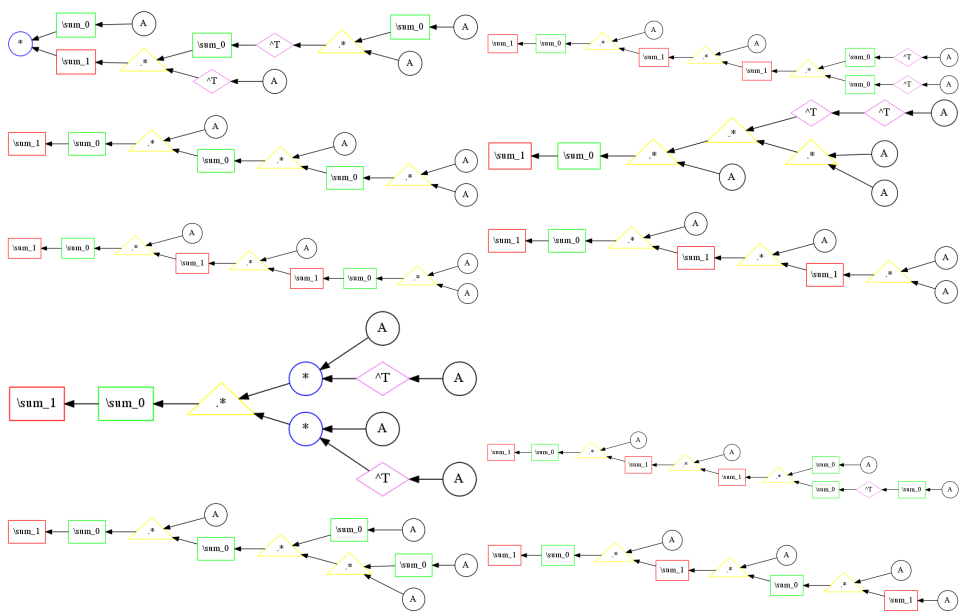

23

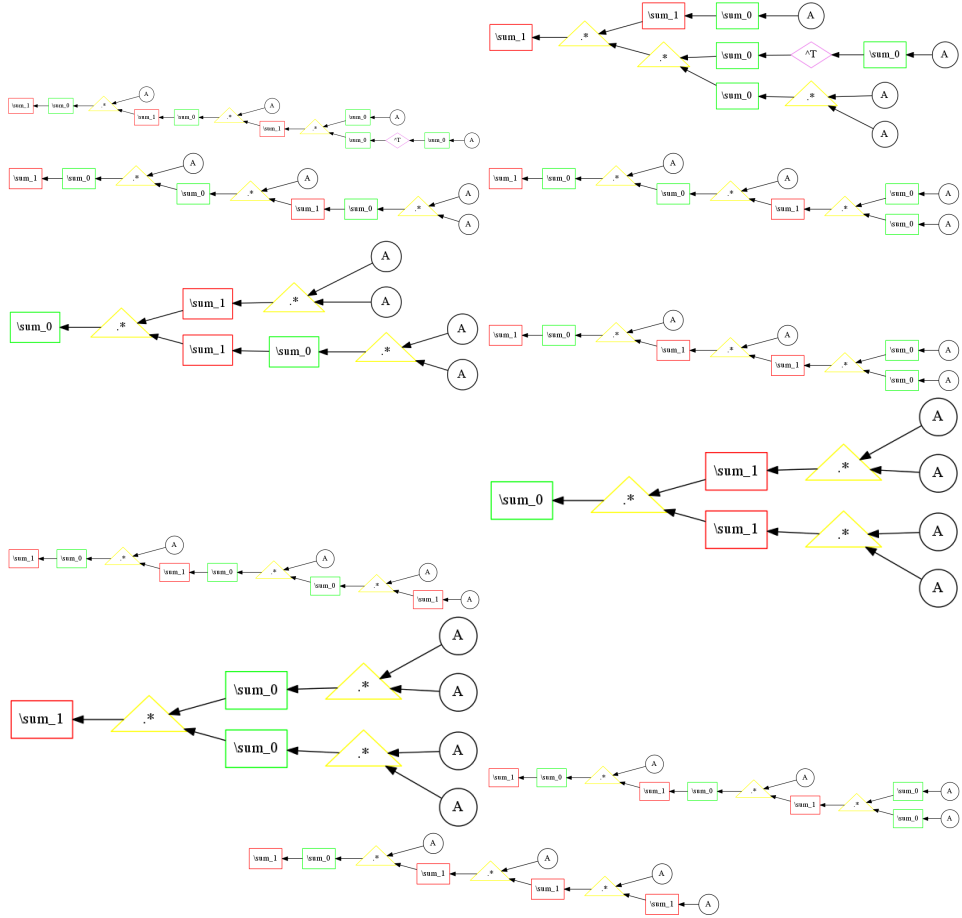

**k = 5**

```
n = 7;
m = 8;
A = randn(n, m);
nset = dec2bin(0:(2^(n) - 1));
mset = dec2bin(0:(2^(m) - 1));
original = 0;
for i = 1:size(nset, 1)
  for j = 1:size(mset, 1)
    v = logical(nset(i, :) - '0');
    h = logical(mset(j, :) - '0');
    original = original + (v * A * h') ^ 5;
  end
end

optimized = 2^(n + m - 11) * (20 * (sum(sum((A .* repmat(sum(sum((A .* repmat(sum(sum((A .* repmat(sum((sum(A, 1)
↪    .* repmat(sum(sum(A, 1)', 1), 1, m)), 2), n, m)), 2), 1, m)), 1), 2), n, m)), 1), 2)) + 60 * (sum(sum
↪    ((A .* repmat(sum(sum((A .* repmat(sum(sum((A .* repmat(sum(sum((A .* A), 1), 2), n, m)), 2), 1, m)), 1),
↪    2), n, m)), 1), 2)) + 60 * (sum(sum((A .* repmat(sum(sum((A .* repmat(sum((sum(A,
↪    1) .* sum(A, 1)), 2), n, m)), 2), 1, m)), 1), 2), n, m)), 1), 2)) + 30 * (sum(sum((A .* repmat(sum((A
↪    .* repmat(sum(sum((A .* repmat(sum((sum(A, 1) .* sum(A, 1)), 2), n, m)), 1), 2), n, m)), 1), n, 1)),
↪    1), 2)) + 20 * (sum(sum((A .* repmat(sum(sum((A .* A), 1), 2), n, m)), 1), 2)) + 240 * (sum((sum((repmat(sum((repmat(sum(A', 1)
↪    , m, 1) .* A'), 2), 1, n) .* A'), 1) .* sum((A * A'), 1)), 2)) + −20 * (sum(sum((A .* repmat(sum(sum
↪    ((A .* repmat(sum((repmat(sum(A, 1), n, 1) .* repmat(sum(A, 1), n, 1) .* A)), 1), n, 1)), 1), 2), n,
↪    m)), 1), 2)) + −60 * (sum(sum((A .* repmat(sum((sum((A .* A), 1) .* sum((A .* A), 1)), 2), n, m)),
↪    1), 2)) + −120 * (sum(sum((A .* repmat(sum(sum((A .* repmat(sum((A .* repmat(sum((A .* A), 1), n, 1))
↪    , 1), n, 1)), 1), 2), n, m)), 1), 2)) + 60 * (sum(sum((A .* repmat(sum(sum((A .*
↪    repmat(sum(sum((A .* A), 1), 2), n, m)), 1), n, 1)), 1), 2), n, m)), 1), 2)) + 120 * (sum(sum((A .*
↪    repmat(sum((A .* repmat(sum((A .* repmat(sum(sum((A .* A), 1), 2), n, m)), 2), 1, m)), 1), n, 1)), 1)
↪    , 2)) + 20 * (sum(sum((A .* repmat(sum(sum((A .* repmat(sum(sum(A, 1) .* sum(A,
↪    1)), 2), n, m)), 1), 2), n, m)), 1), 2)) + 120 * (sum(sum((A .* repmat(sum((A .*
↪    repmat(sum((A .* repmat(sum((sum(A, 1) .* sum(A, 1)), 2), n, m)), 2), 1, m)), 1), n, 1)), 1), 2)) +
↪    120 * (sum(sum((A .* repmat(sum(sum((A .* repmat(sum((sum(A', 1) .* sum(A', 1)), 2), n,
↪    m)), 2), 1, m)), 1), n, 1)), 1), 2)) + 2 * (sum(sum((A .* repmat(sum(sum((A .* repmat(sum((A .*
↪    repmat(sum((sum(A, 1) .* repmat(sum(sum(A, 1)', 1), 1, m)), 2), n, m)), 1), 2), n, m)), 1), 2), n, m)
↪    ), 1), 2)) + −20 * repmat(sum((A .* repmat(sum(((repmat(sum(A', 1), m, 1) .* A') .* repmat(sum(A',
↪    1), m, 1)) .* sum(A', 1)), 2), n, m)), 1), 2)) + −60 * (sum(sum((A .* repmat(sum((sum((A .* A),
↪    2) .* sum((A .* A), 2)), 1), n, m)), 1), 2)) + −120 * (sum(sum((A .* repmat(sum(sum((A .* repmat(sum
↪    ((A .* repmat(sum((A .* A), 2), 1, m)), 2), 1, m)), 1), 2), n, m)), 1), 2)) + 30 * (sum((sum((A .* A)
```

```
                     , 2) .* repmat(sum(sum((A .* repmat(sum(sum((A .* A), 1), 2), n, m)), 1), 2), n, 1)), 1)) + 60 * (sum
                     (sum((A .* repmat(sum(sum(((A * A') .* (A * A')), 1), 2), n, m)), 1), 2)) + 40 * (sum(sum((A .*
                     repmat(sum(sum(((A .* A) .* (A .* A)), 1), 2), n, m)), 1), 2)) + 30 * (sum(sum((A .* repmat(sum(sum((
                     A .* repmat(sum((A .* repmat(sum((sum(A', 1) .* sum(A', 1)), 2), n, m)), 2), 1, m)), 1), 2), n, m)),
                     1), 2)) + 120 * (sum(sum((A .* repmat(sum((A .* repmat(sum((repmat(sum(A, 1), n, 1) .* (repmat(sum(
                     sum(A, 1)', 1), n, m) .* A)), 2), 1, m)), 1), n, 1)), 1), 2)) + −80 * (sum(sum((A .* repmat(sum((A .*
                     repmat(sum((repmat(sum(A, 1), n, 1) .* (repmat(sum(A, 1), n, 1) .* A)), 1), n, 1)), 2), 1, m)), 1),
                     2)) + −80 * ((sum(A, 1) * sum(((repmat(sum(A', 1), m, 1) .* A') .* repmat((sum(A', 1) .* sum(A', 1)),
                     m, 1)), 2))) + −240 * (sum(sum((A .* repmat(sum((A .* repmat(sum((A .* repmat(sum((A .* A), 2), 1, m
                     )), 2), 1, m)), 1), n, 1)), 1), 2) + 120 * (sum(sum((A .* repmat(sum((repmat(sum((repmat(sum(sum(A,
                     1)', 1), n, m) .* A)', 1), n, 1) .* (A * A')), 2), 1, m)), 1), 2)) + −240 * (sum(sum((A .* repmat(sum
                     ((A .* repmat(sum((A .* repmat(sum((A .* A), 1), n, 1)), 1), n, 1)), 2), 1, m)), 1), 2)) + 120 * (sum
                     (sum((A .* repmat(sum((A .* repmat(sum((A .* repmat(sum((sum(A, 1) .* repmat(sum(sum(A, 1)', 1), 1, m
                     )), 2), n, m)), 2), 1, m)), 1), n, 1)), 1), 2)) + 160 * (sum(sum(((repmat(sum(A', 1), m, 1) .* A') .*
                     ((repmat(sum(A, 1), n, 1) .* A)' .* A')), 1), 2)));
normalization = sum(abs(original(:)));
assert(sum(abs(original(:) − optimized(:))) / normalization < 1e−10);
```

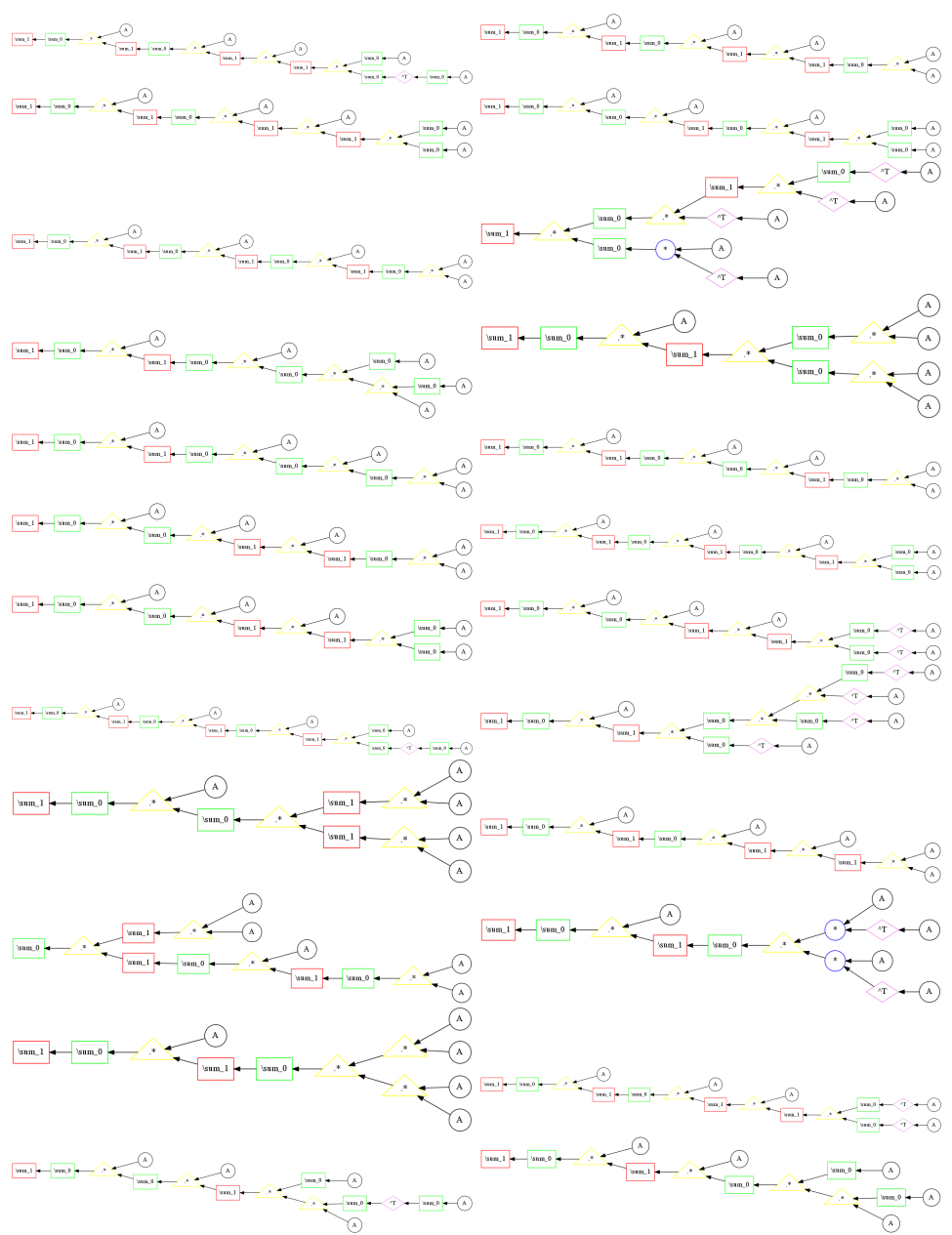

25

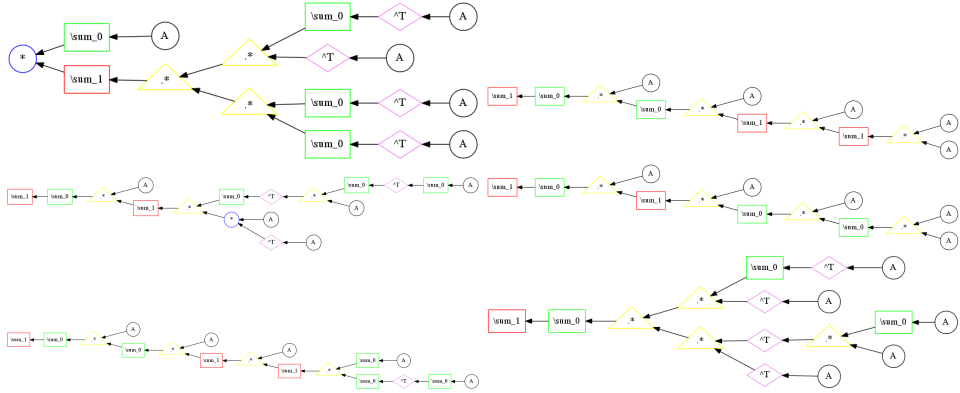

**k = 6**

```
n = 8;
m = 9;
A = randn(n, m);
nset = dec2bin(0:(2^(n) − 1));
mset = dec2bin(0:(2^(m) − 1));
original = 0;
for i = 1:size(nset, 1)
  for j = 1:size(mset, 1)
    v = logical(nset(i, :) − '0');
    h = logical(mset(j, :) − '0');
    original = original + (v * A * h') ^ 6;
  end
end

optimized = 2^(n+m)*(((sum(sum((( A .* A)  .* (( repmat(sum(A, 2), [1, m]) .* repmat(sum(A, 1), [n, 1]))  .* (
          repmat(sum(A, 2), [1, m]) .* repmat(sum(A, 1), [n, 1])))), 2), 1)  .* 360) ...
  + (sum(sum((( A .* repmat(sum(A, 2), [1, m])) .* ( ( A .* repmat(sum(A, 2), [1, m])) .* repmat(sum(( A .* A
          ), 1), [n, 1]))), 2), 1)  .* 360) ...
  + (sum((( sum(A, 1) .* sum(A, 1)) .* ( ( sum(A, 1) .* sum(A, 1)) .* ( sum(A, 1) .* sum(A, 1)))), 2)  .* 16)
          ...
  + ((sum(( sum(A, 1) .* sum(A, 1)), 2)  .* sum(( ( sum(A, 1) .* sum(A, 1)) .* ( sum(A, 1) .* sum(A, 1))), 2))
           .* −30) ...
  + ( ( ( sum(A, 1) * ( (A') * sum(A, 2))) * ( sum(A, 1) * ( (A') * sum(A, 2))))  .* 360) ...
  + ( ( ( sum(A, 1) * (A')) * ( ( A .* ( A .* A)) * (sum(A, 1)')))  .* 480) ...
  + (sum((( sum(A, 2) .* sum(A, 2)) .* ( ( sum(A, 2) .* sum(A, 2)) .* ( sum(A, 2) .* sum(A, 2)))), 1)  .* 16)
          ...
  + (sum(sum((( A .* repmat(sum(A, 1), [n, 1])) .* ( ( A .* repmat(sum(A, 1), [n, 1])) .* repmat(sum(( A .* A
          ), 2), [1, m]))), 2), 1)  .* 360) ...
  + ((sum(( sum(A, 1) .* sum(A, 1)), 2)  .* sum(( ( sum(A, 1) .* sum(A, 1)) .* sum(( A .* A), 1)), 2))  .*
          −180) ...
  + (sum((repmat(sum(sum(A, 2), 1), [n, 1])  .* sum(( ( repmat(sum(A, 2), [1, m]) .* repmat(sum(A, 1), [n, 1])
          ) .* ( A .* ( A .* A))), 2)), 1)  .* 480) ...
  + (sum((repmat(sum(sum(A, 2), 1), [n, 1])  .* sum((sum(A, 2)  .* (sum(A, 2)  .* sum(( ( repmat(sum(A, 2), [1, m
          ]) .* repmat(sum(A, 1), [n, 1])) .* A), 2)))), 1)  .* −240) ...
  + (sum((sum(( A .* repmat(sum(A, 1), [n, 1])), 2)  .* (sum(A, 2)  .* sum(repmat(sum((repmat(sum(A, 2), [1, m
          ]) .* ( repmat(sum(A, 2), [1, m]) .* repmat(sum(A, 1), [n, 1]))), 1), [n, 1]), 2))), 1)  .* 360)
          ...
  + ( ( ( sum(sum(A, 2), 1) * ( sum(A, 1) * (A'))) * ( ( A * (A')) * sum(A, 2)))  .* 720) ...
  + ((( sum(sum(A, 2), 1) * ( sum(sum(A, 2), 1) * sum(sum(A, 2), 1))  .* ( sum(sum(A
          , 2), 1)  .* sum(sum(A, 2), 1))))  .* 1) ...
  + ((sum(( sum(A, 1) .* sum(A, 1)), 2)  .* (( sum(sum(A, 2), 1) * sum(sum(A, 2), 1))  .* ( sum(sum(A, 2), 1)
           .* sum(sum(A, 2), 1))))  .* 15) ...
  + (sum((sum(( A .* repmat(sum(A, 1), [n, 1])), 2)  .* (sum(A, 2)  .* repmat(( sum(sum(A, 2), 1) .* ( sum(sum
          (A, 2), 1) .* sum(sum(A, 2), 1))), [n, 1]))), 1)  .* 120) ...
  + ((sum(( sum(A, 2) .* sum(A, 2)), 1)  .* (( sum(sum(A, 2), 1) * sum(sum(A, 2), 1))  .* ( sum(sum(A, 2), 1)
           .* sum(sum(A, 2), 1))))  .* 15) ...
  + ((sum(sum(( A .* A), 2), 1)  .* (( sum(sum(A, 2), 1) * sum(sum(A, 2), 1))  .* ( sum(sum(A, 2), 1) .* sum(
          sum(A, 2), 1))))  .* 15) ...
  + (sum(sum((repmat(( sum(A, 2) .* sum(A, 2)), [1, m])  .* (( repmat(sum(A, 2), [1, m]) .* repmat(sum(A, 1),
          [n, 1]))  .* repmat(sum(A, 2), [1, m]) .* repmat(sum(A, 1), [n, 1])))), 1), 2)  .* −30) ...
  + (sum((sum((repmat(sum(A, 2), [1, m])  .* ( repmat(sum(A, 2), [1, m]) .* repmat(sum(A, 1), [n, 1]))), 1)
           .* sum((repmat(sum(A, 2), [1, m])  .* ( repmat(sum(A, 2), [1, m]) .* repmat(sum(A, 1), [n, 1]))),
          1)), 2)  .* 45) ...
  + (sum((sum(( A .* repmat(sum(A, 2), [1, m])), 1)  .* (sum(( A .* repmat(sum(A, 2), [1, m])), 1)  .* repmat(
          sum(( sum(A, 1) .* sum(A, 1)), 2), [1, m]))), 2)  .* 180) ...
  + (sum((sum(( A .* repmat(sum(A, 2), [1, m])), 1)  .* (sum(A, 1)  .* sum(( ( repmat(sum(A, 2), [1, m]) .*
          repmat(sum(A, 1), [n, 1])) .* A), 1))), 2)  .* −360) ...
  + ((sum(( sum(A, 1) .* sum(A, 1)), 2)  .* ( sum(( sum(A, 1) .* sum(A, 1)), 2)  .* sum(( sum(A, 1) .* sum(A,
          1)), 2)))  .* 15) ...
  + (sum((sum((repmat(sum(A, 1), [n, 1])  .* ( repmat(sum(A, 2), [1, m]) .* repmat(sum(A, 1), [n, 1]))), 1)
           .* sum((repmat(sum(A, 1), [n, 1])  .* ( repmat(sum(A, 2), [1, m]) .* repmat(sum(A, 1), [n, 1]))),
          1)), 2)  .* −30) ...
  + ((sum(sum(A, 2), 1)  .* (sum(sum(A, 2), 1)  .* ( sum(( sum(A, 1) .* sum(A, 1)), 2)  .* sum(( sum(A, 1) .*
          sum(A, 1)), 2))))  .* 45) ...
  + ((( sum(A, 1) * sum(A, 1)) * ( ( repmat(sum(A, 1), [m, 1]) * (A')) * ( A * (sum(A, 1)'))))  .* 180) ...
  + (sum((sum(( A .* repmat(sum(A, 1), [n, 1])), 2)  .* (( sum(A, 2)  .* repmat(sum((sum(A, 2)  .* sum(A, 2)), 1), [n, 1]))  .*
          sum(repmat(( sum(A, 1) .* sum(A, 1)), [n, 1]), 2))), 1)  .* 360) ...
  + ( ( sum(( A .* A), 1) * ( ( (A') * sum(A, 2))  .* ( (A') * sum(A, 2))))  .* −360) ...
```

26

```
+ (sum((  sum(( A .* A), 1)  .* ( ( sum(A, 1) .* sum(A, 1)) .* ( sum(A, 1) .* sum(A, 1)))), 2)   .* 240) ...
+ ((sum(sum(( A .* A), 2), 1)   .* sum(( ( sum(A, 1) .* sum(A, 1)) .* ( sum(A, 1) .* sum(A, 1))), 2))   .*
    ↪ −30) ...
+ ((sum(( sum(A, 1) .* sum(A, 1)), 2)   .* ( sum(( sum(A, 1) .* sum(A, 1)), 2) .* sum(sum(( A .* A), 2), 1)))
    ↪   .* 45) ...
+ ((sum(sum(A, 2), 1)   .* ( sum(sum(A, 2), 1) .* sum(( ( sum(A, 1) .* sum(A, 1)) .* sum(( A .* A), 1)), 2)))
    ↪   .* −180) ...
+ ((( sum(sum(A, 2), 1) .* sum(sum(A, 2), 1))   .* ( sum(( sum(A, 1) .* sum(A, 1)), 2) .* sum(sum(( A .* A),
    ↪ 2), 1)))   .* 90) ...
+ (sum(sum((( repmat(sum(( A .* A), 2), [1, m]) .* repmat(sum(( A .* A), 1), [n, 1]))   .* ( A .* A)), 2), 1)
    ↪   .* 360) ...
+ (sum((sum(( A .* A), 1)   .* ( sum(( A .* A), 1) .* repmat(sum(( sum(A, 1) .* sum(A, 1)), 2), [1, m]))), 2)
    ↪   .* −90) ...
+ ((sum(sum(( A .* A), 2), 1)   .* sum(( ( sum(A, 2) .* sum(A, 2)) .* ( sum(A, 2) .* sum(A, 2))), 1))   .*
    ↪ −30) ...
+ ((sum(( sum(A, 2) .* sum(A, 2)), 1)   .* ( sum(( sum(A, 2) .* sum(A, 2)), 1) .* sum(sum(( A .* A), 2), 1)))
    ↪   .* 45) ...
+ (sum((sum(( A .* repmat(sum(A, 2), [1, m])), 1)   .* (sum(( A .* repmat(sum(A, 2), [1, m])), 1) .* repmat(
    ↪ sum(sum(( A .* A), 2), 1), [1, m])), 2)   .* 180) ...
+ (sum((sum(( A .* repmat(sum(A, 1), [n, 1])), 2)   .* (sum(( A .* repmat(sum(A, 1), [n, 1])), 2) .* repmat(
    ↪ sum(sum(( A .* A), 2), 1), [n, 1])), 1)   .* 180) ...
+ (sum((( sum(A, 2) .* repmat(sum(sum(A, 2), 1), [n, 1]))   .* (sum(( A .* repmat(sum(A, 1), [n, 1])), 2) .*
    ↪ repmat(sum(sum(( A .* A), 2), 1), [n, 1]))), 1)   .* 360) ...
+ (sum((sum(( A .* repmat(sum(A, 1), [n, 1])), 2)   .* sum(( repmat( sum(A, 1) .* sum(A, 1)), [n, 1]) .* ( A
    ↪   .* repmat(sum(A, 1), [n, 1]))), 2)), 1)   .* −240) ...
+ (sum((A  .* repmat((repmat(sum(A, 2), [1, m]) .* repmat(sum(A, 1), [n, 1])) .* A), 1))), [n, 1])), 2), 1)   .* −240) ...
+ ((sum(( sum(A, 2) .* sum(A, 2)), 1)   .* sum(( ( sum(A, 1) .* sum(A, 1)) .* ( sum(A, 1) .* sum(A, 1))), 2))
    ↪   .* −30) ...
+ ((( sum(sum(A, 2), 1) .* sum(sum(A, 2), 1))   .* sum(( ( sum(A, 2) .* sum(A, 2)) .* sum(( A .* A), 2)), 1))
    ↪   .* −180) ...
+ ((( sum(sum(A, 2), 1) .* sum(sum(A, 2), 1))   .* sum(( ( A .* A) .* ( A .* A)), 2), 1))   .* 60) ...
+ (sum((sum((repmat(sum(A, 2), [1, m])  .* ( repmat(sum(A, 2), [1, m]) .* repmat(sum(A, 1), [n, 1]))), 2)
    ↪   .* sum((repmat(sum(A, 2), [1, m])  .* ( repmat(sum(A, 2), [1, m]) .* repmat(sum(A, 1), [n, 1]))),
    ↪ 2)), 1)   .* −30) ...
+ ((sum(sum(A, 2), 1)   .* sum((sum(repmat(( sum(A, 2) .* sum(A, 2)), [1, m]), 1)   .* sum((repmat(sum(A, 2),
    ↪ [1, m])   .* ( repmat(sum(A, 2), [1, m]) .* repmat(sum(A, 1), [n, 1]))), 1)), 2))   .* 45) ...
+ (sum((sum(( A .* repmat(sum(A, 2), [1, m])), 1)   .* (sum(( A .* repmat(sum(A, 2), [1, m])), 1) .* repmat
    ↪ (( sum(sum(A, 2), 1) .* sum(sum(A, 2), 1)), [1, m]))), 2)   .* 180) ...
+ ((( sum(sum(A, 2), 1) .* sum(sum(A, 2), 1))   .* ( sum(( sum(A, 1) .* sum(A, 1)), 2) .* sum(( sum(A, 2) .*
    ↪ sum(A, 2)), 1)))   .* 90) ...
+ (sum((sum(( A .* repmat(sum(A, 1), [n, 1])), 2)   .* (sum(( A .* repmat(sum(A, 1), [n, 1])), 2) .* repmat
    ↪ (( sum(sum(A, 2), 1) .* sum(sum(A, 2), 1)), [n, 1])), 1)   .* 180) ...
+ ( (  ( sum(A, 1) * ( (A') * A) ) * (  ( (A') * A) .* sum(A, 1)'))) .* 360) ...
+ (sum((sum(( A .* repmat(sum(A, 1), [n, 1])), 2)   .* (sum(( A .* repmat(sum(A, 1), [n, 1])), 2) .* repmat(
    ↪ sum(( sum(A, 2) .* sum(A, 2)), 1), [n, 1])), 1)   .* 180) ...
+ ((sum(( sum(A, 1) .* sum(A, 1)), 2)   .* ( sum(( sum(A, 1) .* sum(A, 1)), 2) .* sum(( sum(A, 2) .* sum(A,
    ↪ 2)), 1)))   .* 45) ...
+ ( ( ( (sum(A, 2)') * A) * ( (A') * A) .* ((A') * sum(A, 2)))) .* 360) ...
+ (sum((sum(( A .* repmat(sum(A, 2), [1, m])), 1)   .* sum(repmat(( sum(A, 2) .* sum(A, 2)), [1, m]) .* ( A
    ↪   .* repmat(sum(A, 2), [1, m]))), 1)), 2)   .* −240) ...
+ ((( (sum(A, 2)') * A)   * ( (A') * ( repmat(sum(A, 2), [1, n]) * ( sum(A, 2) .* sum(A, 2))))) .* 180) ...
+ ( ( ( (sum(A, 2)') * A) * ( ( A .* ( A .* A))') * sum(A, 2))) .* 480) ...
+ ((sum(( sum(A, 2) .* sum(A, 2)), 1)   .* sum(( ( sum(A, 2) .* sum(A, 2)) .* ( sum(A, 2) .* sum(A, 2))), 1))
    ↪   .* −30) ...
+ ((sum(( sum(A, 2) .* sum(A, 2)), 1)   .* ( sum(( sum(A, 2) .* sum(A, 2)), 1) .* sum(( sum(A, 2) .* sum(A,
    ↪ 2)), 1)))   .* 15) ...
+ ((sum(( sum(A, 2) .* sum(A, 2)), 1)   .* sum(( ( sum(A, 2) .* sum(A, 2)) .* sum(( A .* A), 2)), 1))   .*
    ↪ −180) ...
+ (sum((( sum(( A .* A), 2) .* sum(( A .* A), 2))   .* repmat(sum(( sum(A, 2) .* sum(A, 2)), 1), [n, 1])), 1)
    ↪   .* −90) ...
+ ((sum(( sum(A, 1) .* sum(A, 1)), 2)   .* ( sum(( sum(A, 2) .* sum(A, 2)), 1) .* sum(sum(( A .* A), 2), 1)))
    ↪   .* 90) ...
+ ( sum(( ( sum(A, 1) .* sum(A, 1)) .* sum(( ( A .* A) .* ( A .* A)), 1)), 2) .* −480) ...
+ (( sum(( sum(A, 1) .* sum(A, 1)), 2) .* sum(sum(( ( A .* A) .* ( A .* A)), 2), 1))  .* 60) ...
+ ( sum(( sum(( A .* A), 1) .* sum(( ( A .* A) .* ( A .* A)), 1)), 2) .* −480) ...
+ (sum((sum(repmat(( sum(A, 2) .* sum(A, 2)), [1, m]), 1)   .* sum(( ( A .* A) .* ( A .* A)), 1)), 2)   .* 60)
    ↪ ...
+ ( ( ( sum(A, 1) .* sum(( A .* A), 1)) * ( ( (A') * A) * (sum(A, 1)'))) .* −720) ...
+ ((sum(sum(A, 2), 1)   .* sum((( ( repmat(sum(A, 2), [1, m]) .* repmat(sum(A, 1), [n, 1])) .* A)  .*
    ↪ repmat(sum(( A .* A), 1), [n, 1])), 2), 1)   .* −720) ...
+ ((sum(( sum(A, 2) .* sum(A, 2)), 1)   .* sum(( ( sum(A, 1) .* sum(A, 1)) .* sum(( A .* A), 1)), 2))   .*
    ↪ −180) ...
+ ((sum(sum(( A .* A), 2), 1)   .* sum(( ( sum(A, 1) .* sum(A, 1)) .* sum(( A .* A), 1)), 2))   .* −180) ...
+ (sum(( ( sum(A, 1) .* sum(A, 1)) .* ( sum(( A .* A), 1) .* sum(( A .* A), 1))), 2)   .* 720) ...
+ ( sum(( sum(( A .* A), 1) .* ( sum(A, 1) .* sum(( A .* A), 1))), 2)   .* 240) ...
+ (sum(sum((repmat(sum(( A .* A), 1), [n, 1]) .* (repmat(sum(sum(A, 2), 1), [n, 1])),
    ↪ [1, m])   .* repmat(sum(( A .* A), 1), [n, 1])), 2), 1)   .* −90) ...
+ (sum((sum(repmat(( sum(A, 2) .* sum(A, 2)), [1, m]), 1)   .* ( sum(( A .* A), 1) .* sum(( A .* A), 1))), 2)
    ↪   .* −90) ...
+ (sum((sum(( A .* A), 2)   .* ( sum(( A .* A), 2) .* repmat(( sum(sum(A, 2), 1) .* sum(sum(A, 2), 1)), [n,
    ↪ 1]))), 1)   .* −90) ...
+ ((sum(( sum(A, 1) .* sum(A, 1)), 2)   .* ( sum(sum(( A .* A), 2), 1) .* sum(sum(( A .* A), 2), 1)))   .* 45)
    ↪ ...
+ ((( sum(sum(A, 2), 1) .* sum(sum(A, 2), 1))   .* ( sum(sum(( A .* A), 2), 1) .* sum(sum(( A .* A), 2), 1)))
    ↪   .* 45) ...
+ ((( sum(sum(A, 2), 1) .* sum(sum(A, 2), 1))   .* ( sum(( sum(A, 2) .* sum(A, 2)), 1) .* sum(sum(( A .* A),
    ↪ 2), 1)))   .* 90) ...
+ (sum(( sum(( A .* A), 2) .* ( ( sum(A, 2) .* sum(A, 2)) .* ( sum(A, 2) .* sum(A, 2)))), 1)   .* 240) ...
+ ( sum(( ( sum(A, 2) .* sum(A, 2)) .* sum(( ( A .* A) .* ( A .* A)), 2)), 1)   .* −480) ...
+ (sum(( ( sum(A, 2) .* sum(A, 2)) .* ( sum(( A .* A), 2) .* sum(( A .* A), 2))), 1)   .* 720) ...
```

27

```
       + ((sum(sum(A, 2), 1)  .* sum((sum(( A .* A), 2)  .* sum(( ( repmat(sum(A, 2), [1, m]) .* repmat(sum(A, 1),
↪         [n, 1])) .* A), 2)), 1))  .* −720) ...
       + ( ( ( (sum(A, 2)') * ( A * (A'))) * ( sum(A, 2) .* sum(( A .* A), 2))) .* −720) ...
       + ( sum(( sum(( A .* A), 2) .* sum(( ( A .* A) .* ( A .* A)), 2)), 1) .* −480) ...
       + ( sum(( sum(( A .* A), 2) .* ( sum(( A .* A), 2) .* sum(( A .* A), 2))), 1) .* 240) ...
       + ((sum(( sum(A, 1) .* sum(A, 1)), 2)  .* sum(( ( sum(A, 2) .* sum(A, 2)) .* sum(( A .* A), 2)), 1))  .*
↪         −180) ...
       + ((sum(( sum(( A .* A), 2), 1)  .* sum(( ( sum(A, 2) .* sum(A, 2)) .* sum(( A .* A), 2)), 1))  .* −180) ...
       + (sum((sum(( A .* repmat(sum(A, 1), [n, 1])), 2)  .* (sum(A, 2)  .* sum(( ( repmat(sum(A, 2), [1, m]) .*
↪         repmat(sum(A, 1), [n, 1])) .* A), 2))), 1)  .* −360) ...
       + ( ( ( ( sum(A, 1) * (A')) .* ( sum(A, 1) * (A'))) * sum(( A .* A), 2))  .* −360) ...
       + (sum((sum(repmat(( sum(A, 1) .* sum(A, 1)), [n, 1]), 2)  .* ( sum(( A .* A), 2) .* sum(( A .* A), 2))), 1)
↪         .* −90) ...
       + ((sum(( sum(A, 2) .* sum(A, 2)), 1)  .* ( sum(sum(( A .* A), 2), 1) .* sum(sum(( A .* A), 2), 1)))  .* 45)
↪         ...
       + ( sum(sum(( ( A .* A) .* ( ( A .* A) .* ( A .* A))), 2), 1) .* 256) ...
       + ( ( sum(sum(( A .* A), 2), 1) .* sum(sum(( ( A .* A) .* ( A .* A)), 2), 1)) .* 60) ...
       + ((sum(sum(( A .* A), 2), 1)  .* ( sum(sum(( A .* A), 2), 1) .* sum(sum(( A .* A), 2), 1)))  .* 15) ...
       + (sum(sum((repmat(sum(( A .* A), 1), [n, 1])  .* ( repmat(sum(( A .* A), 2), [1, m]) .* repmat(sum(( A .* A
↪         ), 1), [n, 1]))), 1), 2)  .* −90) ...
       + ((repmat(sum(sum(A, 2), 1), [1, m])  * ( ( ( (A') * A) .* ( (A') * A)) * repmat(sum(sum(A, 2), 1), [m, 1]
↪         ))  .* 90) ...
       + (sum(sum((repmat(sum(( A .* A), 2), [1, m])  .* ( repmat(sum(( A .* A), 2), [1, m]) .* repmat(sum(( A .* A
↪         ), 1), [n, 1]))), 2), 1)  .* −90) ...
       + ( sum(( ( ( A * (A')) .* ( A * (A'))) * ( sum(A, 2) .* sum(A, 2))), 1) .* −360) ...
       + ( sum(( ( ( A * (A')) .* ( A * (A'))) * sum(( A .* A), 2)), 1) .* −360) ...
       + (sum(( ( ( A * (A')) .* ( A * (A'))) * sum(repmat(( sum(A, 1) .* sum(A, 1)), [n, 1]), 2)), 1)  .* 90) ...
       + (sum(( ( ( A * (A')) .* ( A * (A'))) * repmat(sum(( sum(A, 2) .* sum(A, 2)), 1), [n, 1])), 1)  .* 90) ...
       + ( sum(( sum(A, 1) .* sum(A, 1)) * ( ( (A') * A) .* ( (A') * A))), 2) .* −360) ...
       + ( sum(( sum(( A .* A), 1) * ( ( (A') * A) .* ( (A') * A))), 2) .* −360) ...
       + (( repmat(sum(sum(( A .* A), 2), 1), [1, m]) * sum(( ( (A') * A) .* ( (A') * A)), 2))  .* 90) ...
       + ( sum(sum(( A .* ( ( A * (A')) * ( A .* ( A .* A)))), 2), 1) .* 480) + ( sum(sum(( ( ( A * (A')) * A) .*
↪         ( ( A * (A')) * A)), 2), 1)  .* 120))) / 4096;
normalization = sum(abs(original(:)));
assert(sum(abs(original(:) − optimized(:))) / normalization < 1e−10);
```

28



\end{document}